\documentclass[a4paper,fleqn]{cas-dc}
\usepackage{amsmath}
\usepackage{amsthm}
\usepackage[authoryear,longnamesfirst]{natbib}
\newtheorem{theorem}{Theorem}

\newtheorem{corollary}[theorem]{Corollary}

\newtheorem{definition}{Definition}
\newtheorem{property}{Property}
\usepackage{xcolor}
\usepackage{colortbl}
\definecolor{secondcolor}{RGB}{189,215,238}
\definecolor{firstcolor}{RGB}{255,153,153}

\usepackage{algorithm}
\usepackage{algorithmic}
\newcommand{\greytextbf}[1]{\cellcolor{gray!25}\textbf{#1}}
\def\tsc#1{\csdef{#1}{\textsc{\lowercase{#1}}\xspace}}
\tsc{WGM}
\tsc{QE}
\tsc{EP}
\tsc{PMS}
\tsc{BEC}
\tsc{DE}

\begin{document}

\let\WriteBookmarks\relax
\def\floatpagepagefraction{1}
\def\textpagefraction{.001}

\shorttitle{Neural Networks}

\shortauthors{Y. Huang et~al.}

\title [mode = title]{CogniSNN: Enabling Neuron-Expandability, Pathway-Reusability, and Dynamic-Configurability with Random Graph Architectures in Spiking Neural Networks}

%
\cortext[cor1]{Corresponding author.}

\author[1,2]{Yongsheng Huang}[type=editor,
                        auid=000,bioid=1,
                        orcid=0009-0001-6620-4343]
\credit{Writing - original draft, Conceptualization, Data curation, Formal analysis, Investigation, Methodology, Software, Validation, Visualization}

\author[1]{Peibo Duan}[style=chinese]
\cormark[1] 
\ead{duanpeibo@swc.neu.edu.cn}
\credit{Writing - review $\&$ editing, Conceptualization, Funding acquisition, Investigation, Methodology, Project administration, Resources, Supervision}

\author[3]{Yujie Wu}[style=chinese]
\credit{Writing - review $\&$ editing, Conceptualization, Formal analysis, Methodology, Project administration, Resources, Supervision}

\author[4]{Kai Sun}[style=chinese]
\credit{Writing - original draft, Data curation, Software, Validation}

\author[1]{Zhipeng Liu}[style=chinese]
\credit{Writing - original draft, Data curation, Software, Visualization}
\author[1]{Changsheng Zhang}[style=chinese]
\credit{Writing - review $\&$ editing, Project administration, Resources}
\author[1]{Bin Zhang}[style=chinese]
\credit{Writing - review $\&$ editing, Project administration, Funding acquisition, Resources}

\author[2]{Mingkun Xu}[style=chinese]
\cormark[1] 
\ead{xumingkun@gdiist.cn} 
\credit{Writing - review $\&$ editing, Funding acquisition, Methodology, Project administration, Resources, Supervision}

\affiliation[1]{organization={School of Software},
    addressline={Northeastern University}, 
    city={Shenyang},
    postcode={110000}, 
    country={China}
    }


\affiliation[2]{organization={Guangdong Institute of Intelligence Science and Technology},
    city={Zhuhai},
    postcode={519000}, 
    country={China}
    }


\affiliation[3]{organization={Department of Computing},
    addressline={The Hong Kong Polytechnic University}, 
    city={Hongkong},
    postcode={000000}, 
    country={China}}

\affiliation[4]{organization={Department of Data Science and AI},
    addressline={Monash University}, 
    city={Melbourne},
    postcode={3000}, 
    country={Australia}}

\begin{abstract}
    Spiking neural networks (SNNs), regarded as the third generation of artificial neural networks, are expected to bridge the gap between artificial intelligence and computational neuroscience. However, most mainstream SNN research directly adopts the rigid, chain-like hierarchical architecture of traditional artificial neural networks (ANNs), ignoring key structural characteristics of the brain. Biological neurons are stochastically interconnected, forming complex neural pathways that exhibit Neuron-Expandability, Pathway-Reusability, and Dynamic-Configurability. In this paper, we introduce a new SNN paradigm, named Cognition-aware SNN (CogniSNN), by incorporating Random Graph Architecture (RGA). Furthermore, we address the issues of network degradation and dimensional mismatch in deep pathways by introducing an improved pure spiking residual mechanism alongside an adaptive pooling strategy. Then, we design a Key Pathway-based Learning without Forgetting (KP-LwF) approach, which selectively reuses critical neural pathways while retaining historical knowledge, enabling efficient multi-task transfer. Finally, we propose a Dynamic Growth Learning (DGL) algorithm that allows neurons and synapses to grow dynamically along the internal temporal dimension. Extensive experiments demonstrate that CogniSNN achieves performance comparable to, or even surpassing, current state-of-the-art SNNs on neuromorphic datasets and Tiny-ImageNet. The Pathway-Reusability enhances the network's continuous learning capability across different scenarios, while the dynamic growth algorithm improves robustness against interference and mitigates the fixed-timestep constraints during neuromorphic chip deployment. This work demonstrates the potential of SNNs with random graph structures in advancing brain-inspired intelligence and lays the foundation for their practical application on neuromorphic hardware. The code is available at \url{https://github.com/Yongsheng124/CogniSNN}.
\end{abstract}

\begin{keywords}
Spiking neural networks \sep Spiking residual learning \sep Random graph theory  \sep Robustness \sep Neuromorphic object recognition
\end{keywords}

\maketitle

\section{Introduction}

Originally envisioned to simulate biological firing processes, Spiking Neural Networks (SNNs), owing to their event-driven nature, ultra-low energy consumption, and rich spatio-temporal dynamics, have garnered significant attention in recent years~\cite{2022_wu_brain}. However, in a relentless pursuit of performance metrics~\citep{2020_deng_rethinking}, exemplified by direct training approaches~\citep{2019_wu_direct, 2021_zheng_going}, current mainstream SNNs predominantly adopt architectures derived from traditional Artificial Neural Networks (ANNs), such as Spiking ResNet ~\citep{2021_hu_spiking}, Spiking  Transformer~\citep{2025_lu_estsformer}, and Spiking Mamba~\citep{2024_li_spikemba}, thereby increasingly deviating from these brain-inspired origins. While these equivalents often achieve performance parity with ANNs in static tasks~\citep{2020_he_comparing}, they fall short of the expectations placed on SNNs as the intersection of computational neuroscience and artificial intelligence toward Artificial General Intelligence (AGI)~\citep{2020_deng_tianjic,2023_xu_unified}. Specifically, existing architectures excel at single-task processing, fueling applications like autonomous driving and ChatGPT, but struggle significantly when faced with real-world, multi-task scenarios~\citep{2020_tyagi_challenges}. They suffer from catastrophic forgetting and exhibit weak robustness against interference, capabilities where the biological brain far surpasses artificial systems. These challenges necessitate a revisiting of the structural principles of the brain to explore novel paradigms for SNN design.

The brain consists of a vast number of neurons with stochastic connections, and the connectivity can be abstracted as a Random Graph Architecture (RGA) with small-world properties~\citep{2009_bullmore_complex}.  However, most models employ rigid, chain-like hierarchical architectures ~\citep{2019_xie_exploring}, which fail to reflect the complex topology of biological networks.  While prior works~\citep{2019_xie_exploring, 2024_yan_sampling} have utilized random graphs for Network Architecture Search (NAS), they primarily view the random structure as a search space rather than a functional feature. Other studies~\citep{2021_xu_exploiting,2023_xu_exploiting} introducing graph learning to SNNs focus on processing graph data rather than innovating the network topology itself. Consequently, the potential of RGA to model the intrinsic characteristics of the brain remains largely unexplored.  

Structurally, an RGA consists of three fundamental elements: nodes, paths, and subgraphs. Drawing analogies from these components, we investigate three corresponding phenomena inherent to the biological brain. Specifically, hundreds of millions of stochastically interconnected neurons in the brain form a vast network of neural pathways. To manage highly complex cognitive and behavioral tasks, the brain relies on three core mechanisms: \textbf{Neuron-Expandability}, where the massive scale and depth of neurons and pathways enable the processing of intricate information~\citep{2012_dicarlo_does}; \textbf{Pathway-Reusability}, characterized by functional orthogonality where specific pathways are selectively activated to facilitate continual learning~\citep{2023_han_adaptive}; and \textbf{Dynamic-Configurability}, which involves the continuous growth and apoptosis of synaptic connections to maintain a dynamically reconfigurable state for high adaptability and robustness~\citep{2025_kudithipudi_neuromorphic}. We posit that effectively modeling these three mechanisms by RGA within SNNs  represents a critical breakthrough toward advancing brain-inspired intelligence.

Although RGA provides abundant neural pathways, directly implementing the three aforementioned brain mechanisms presents significant hurdles. First, regarding Neuron-Expandability, deepening random networks leads to degradation and spatial dimension mismatches. Existing residual solutions often introduce floating-point additions, contradicting the spike-based philosophy of SNNs~\citep{2024_zhou_direct}. To address this, we propose a purely spiking OR Gate residual mechanism combined with an Adaptive Pooling scheme, solving the unbounded value accumulation problem while maintaining event-driven computation.

Second, for Pathway-Reusability, current continual learning approaches face limitations. Methods based on architectural adjustments~\citep{2023_han_enhancing,2024_xu_adaptive,2025_shen_context,2025_shi_hybrid} mainly focus on intra-layer structures, failing to apply directly to neural pathways. Meanwhile, regularization-based methods~\citep{2017_li_learning} typically fine-tune the entire network parameters, lacking the granularity to adaptively select specific topological pathways for new knowledge acquisition. To bridge this gap, we introduce Pathway Betweenness Centrality to identify Key Pathways, enabling a novel algorithm,  which adapts to diverse tasks by selectively reusing topological structures.

Third, regarding Dynamic-Configurability, most existing attempts rely on pruning-based methods~\citep{2023_han_enhancing,2025_ma_efficient} that only operate after each epoch within training loops, iteratively removing less influential nodes or weights. This selection and pseudo-dynamic mechanism significantly prolongs training time and fails to capture the dynamic structure plasticity observed in the brain's internal temporal dimension. We therefore propose a Dynamic Growth Learning (DGL) algorithm that simulates the evolution of neural pathways along the temporal dimension, enhancing robustness and hardware deployment flexibility~\citep{2025_yu_efficient} without the high computational overhead of iterative pruning.

Building upon the Cognition-aware SNN (CogniSNN) proposed in our preliminary work~\citep{2025_huang_cognisnn}, we extend the framework by integrating a novel dynamic growth mechanism to achieve Dynamic-Configurability. In addition to this structural evolution, this paper also provides a comprehensive theoretical analysis and extensive comparative experiments. The specific contributions are as follows:

\begin{itemize}
    \item \textbf{Neuron-Expandability}: We propose the OR Gate residual mechanism and Adaptive Pooling. Theoretical analysis and experiments prove that this approach resolves network degradation and the unbounded value accumulation problem, achieving a mathematically strict, pure spiking identity mapping.
    \item  \textbf{Pathway-Reusability}: We define key Pathways using graph theory and implement the Key Pathway-based Learning without Forgetting (KP-LwF) algorithm. This allows the model to adapt to both similar and dissimilar task scenarios by selectively recruiting high-BC or low-BC pathways, mirroring the brain's functional allocation.
    \item \textbf{Dynamic-Configurability}: We implement a dynamic growth mechanism where the active subgraph evolves along the temporal dimension. This significantly enhances model robustness against noise and frame loss while improving deployment flexibility on neuromorphic hardware.
    \item Extensive experiments on three neuromorphic datasets and one static dataset demonstrate that our model, CogniSNN, achieves results comparable to current state-of-the-art models. Crucially, we demonstrate that CogniSNN possesses superior anti-interference capabilities and continual learning potential that traditional chain-like architectures lack.
\end{itemize}
We emphasize that the primary motivation of this paper is to integrate random graph structures into SNNs to model biological phenomena and strengthen the link between neuroscience and artificial intelligence. Rather than relying on complex, pre-trained backbones to chase marginal accuracy gains, we demonstrate that a structurally bio-plausible design, using only vanilla convolutional layers, can yield a system that is inherently robust, adaptable, and capable of lifelong learning.

\section{Related Work}
\subsection{Random Graph Architecture Modeling}
The predominant architectures of SNNs continue to adopt chain-like hierarchical wiring and manually designed connectivity patterns~\citep{2017_xie_aggregated,2019_xie_exploring}, diverging significantly from biological structure of the human brain, which has been confirmed to adhere to an RGA. As early as the 1940s, ~\citet{2004_turing_intelligent} regarded randomly connected unorganized machines as analogous to an infant's cerebral cortex, thereby establishing the first link between computational models and neuroscience. Subsequently, ~\citet{1958_rosenblatt_perceptron} proposed the view that at the birth of each organism, the construction of its nervous system is largely random and unique. ~\citet{1998_watts_collective, 2011_varshney_structural} further discovered that the nervous system of the nematode forms a random graph with small-world properties composed of approximately 300 neurons, providing additional evidence for the inherent randomness of biological nervous systems. These findings in neuroscience prompt us to reconsider the SNN paradigm through the lens of RGA. ~\citet{2019_xie_exploring} proposed replacing the chain connections in traditional ANN paradigms with random graphs, employing extensive random graph sampling for NAS, with results comparable to those of manually designed backbone networks. ~\citet{2024_yan_sampling} later applied this approach to SNNs with similar success.  \citet{2025_carboni_exploring} analyzed trained ANNs from a graph theory perspective, demonstrating that nodes with specific topologies play a decisive role in the model's resilience against catastrophic forgetting. However, these studies have not yet thoroughly explored the potential of random graph topologies to model the complex characteristics of biological brains within SNNs.

\subsection{Residual Structures in SNNs}
\label{sec:ResInSNN}
Mirroring the development trajectory of ANNs, SNNs also encounter the challenge of network degradation caused by excessive depth. Inspired by ResNet~\citep{2016_he_deep}, researchers have explored various residual mechanisms for SNNs. ~\citet{2021_hu_spiking} first introduced skip connections by converting pre-trained ANNs with residual structures into SNNs. However, due to the activation-after-summation design, the benefits of skip connections are diminished by spiking neurons, causing degradation to emerge at depths as shallow as 18 layers. To address this, ~\citet{2021_fang_deep} proposed SEW-ResNet, which utilizes a summation-after-activation structure in SNNs trained directly, effectively scaling networks to 150 layers. More recently, ~\citet{2024_hu_advancing} introduced MS-ResNet, which performs spike activation before convolution, significantly reducing floating-point computations and energy consumption. ~\citet{2025_li_rethinking} proposed densely additive connections, adding extra residual branches between modules to preserve information flow. However, these methods all rely on additive connections between identity and residual paths, leading to the generation of floating-point values. These values propagate through the entire network via identity branches, resulting in non-spiking computation and unbounded value accumulation issues. Furthermore, spikes represent discrete events, whose direct arithmetic summation lacks biological meaning.  Although SEW-ResNet explored alternative operations such as AND (logical multiplication) and IAND (inverted logical multiplication), their operational principles are biologically counterintuitive and yield suboptimal performance. Therefore, we propose introducing a more biologically plausible spiking residual connection.

\section{Preliminaries}
\subsection{Spiking Neuron and Surrogate Gradient}
The most distinctive feature of SNNs is that they employ spiking neurons as basic computational units. Over $T$ discrete timesteps, each spiking neuron integrates its membrane potential and subsequently generates spikes. Most current works adopt the leaky integrate-and-fire (LIF) neuron model. Its spatio-temporal dynamics can be formally described using discrete equations:
\begin{align}
    U_i[t] = \tau \cdot U_i[t-1] +\sum_{j}W_{ij}S_{j}[t] - S_{i}[t-1]\cdot U_{th},\label{Eq1.membrane potential}
\end{align}
\begin{align}
S_i[t]=\Theta\left(U_i[t]-U_{th}\right)=\left\{\begin{array}{ll}
1, & U_i[t] \geq U_{th} \\
0, & U_i[t]< U_{th}
\end{array} \right. ,\label{Eq2.firing}
\end{align}%

\noindent where $U_i[t]$ is the membrane potential of neuron $i$ at time $t$, $\tau$ refers to the decay factor, $W_{ij}$ represents the synaptic weight connecting the pre-synaptic neuron $j$ and neuron $i$, $S_j[t]$ denotes the spike fired by neuron $j$ at time $t$, $U_{th}$ is the firing threshold, and $\Theta$ denotes the Heaviside step function. Here, we employ the soft reset, where the membrane potential is subtracted by the threshold value after firing.

To perform backpropagation on SNNs using the Backpropagation Through Time (BPTT)~\citep{2002_werbos_backpropagation} algorithm,  originally designed for RNNs, the derivative of loss $\mathcal{L}$ with respect to $W_{ij}$ in layer $i$ can be described as:
\begin{align}
    \frac{ \partial \mathcal{L}}{\partial W_{ij}} = \sum_{t=1}^{T}  \frac{ \partial \mathcal{L}}{\partial S_{i}[t]} \frac{ \partial S_{i}[t]}{\partial U_i[t]}  \frac{ \partial U_i[t]}{\partial W_{ij}}. \label{Eq3.bptt}
\end{align}
However, due to the non-differentiability of the Heaviside step function $\Theta$, the term $\frac{ \partial S_{i}[t]}{\partial U_i[t]}$ is undefined. Therefore,~\citet{2018_wu_spatio} proposed using an approximate function to substitute the gradient. Specifically, during backpropagation, the gradient $\sigma'$ of the approximation function $\sigma(x,\alpha)$ is used to surrogate the gradient of $\Theta$; in forward propagation, $\Theta$ is still maintained for generating spikes. The approximation function $\sigma(x,\alpha)$ can be expressed as:
\begin{align}
    \sigma (x,\alpha )=\frac{1}{1+e^{-\alpha x}}. \label{Eq4.sigma}
\end{align}%
In this paper, we use the LIF neuron, and $\alpha$ is set to 4.

\subsection{Betweenness Centrality of Node and Edge}
\label{sec:bc}
In graph theory, betweenness centrality (BC) is a prominent metric used to quantify the centrality and importance of nodes and edges based on shortest paths within an information graph network~\citep{1994_white_betweenness}. Nodes or edges exhibiting higher betweenness centrality typically function as critical intermediaries, facilitating efficient information transfer and exerting significant influence over the global connectivity of the graph.

Formally, for a node $v$, its betweenness centrality $BC(v)$, as defined by~\citet{1977_freeman_set}, is expressed as:
\begin{align}
BC(v) = \sum_{s\ne v\ne t \in V} \frac{
\phi_{st}(v) }{
\phi_{st}}, \label{Eq5.node BC}
\end{align}%
\noindent where $s$ and $t$ denote the source and target nodes, respectively; $\phi_{st}$ denotes the total number of shortest paths from $s$ to $t$, and $\phi_{st}(v)$ represents the number of those shortest paths passing through the node $v$ from $s$ to $t$.

Similarly, the edge betweenness centrality $BC(e)$, introduced by ~\citet  {2002_girvan_community}, is calculated as:
\begin{align}
    BC(e) = \sum_{s\ne t \in V} \frac{\phi_{st}(e) }{\phi_{st}}, \label{Eq6.edge BC}
\end{align}%

\noindent where $\phi_{st}(e)$ denotes the number of the shortest paths between $s$ and $t$ that traverse edge $e$.

\section{Methods}
This study utilizes WS and ER paradigm-driven random graphs to construct an RGA-based SNN, which constitutes the core of our proposed framework, CogniSNN. To achieve Neuron-Expandability, we propose the ResNode along with an Adaptive Pooling strategy. Subsequently, we explore CogniSNN's capability to adapt to new tasks through a Key Pathway-based Learning without Forgetting (KP-LwF) algorithm. Finally, we implement dynamic growth of neural pathways over internal timesteps, thereby endowing the network with enhanced robustness and hardware friendliness.

\begin{figure*}[t]
    \centering
    \includegraphics[width=\textwidth]{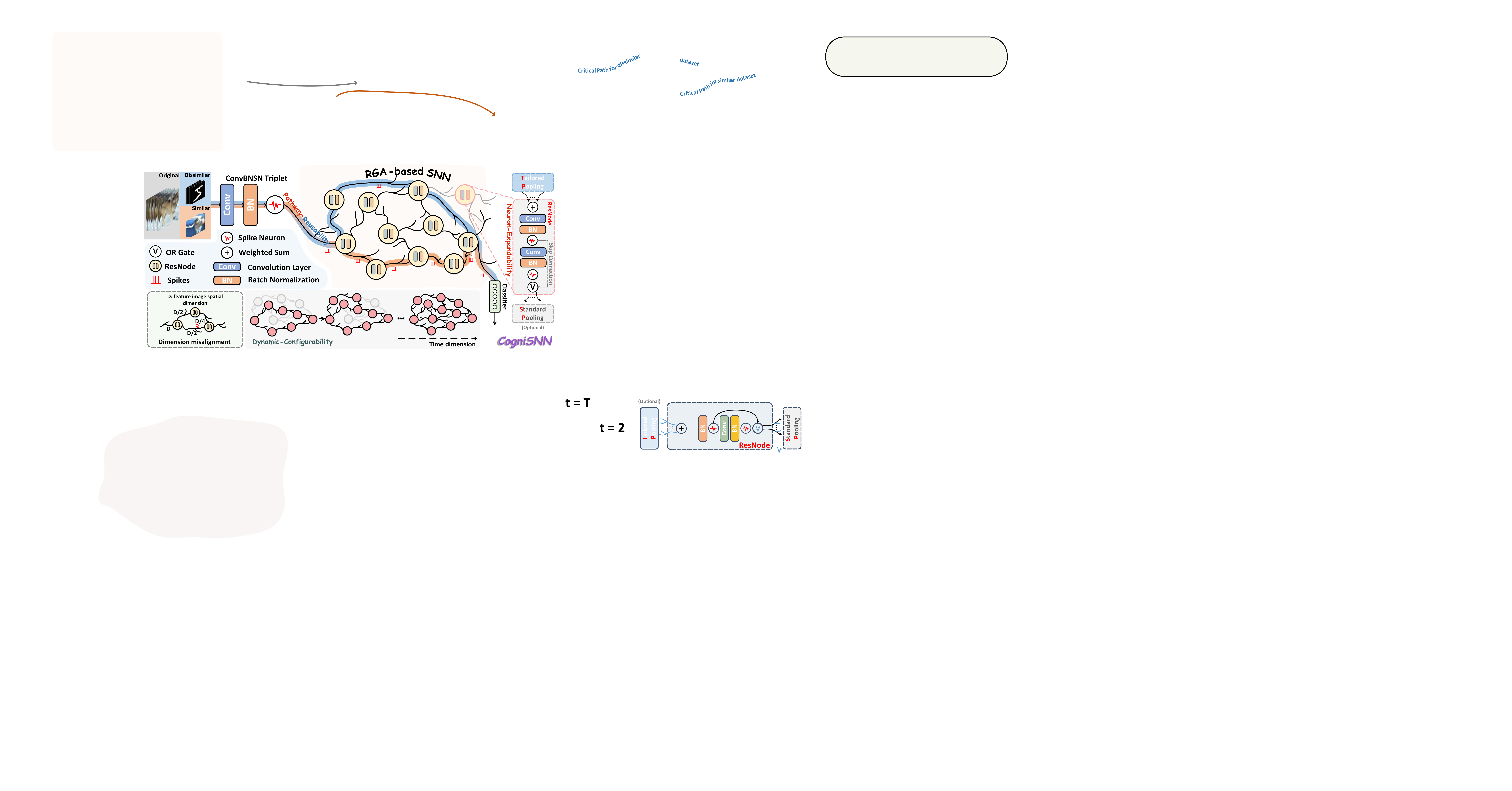}
    \caption{The comprehensive overview of CogniSNN. Spikes propagate from the initial ConvBNSN triplet through the RGA-based backbone to the final classifier. (Right $\&$ Bottom Left) The ResNode structure and the Adaptive Pooling strategy, addressing dimension misalignment, collectively enable Neuron-Expandability. (Center) Regarding Pathway-Reusability, high-BC pathways (orange) and low-BC pathways (blue) are highlighted, corresponding to adaptation in similar and dissimilar scenarios, respectively. (Bottom) The graph topology progressively expands along the temporal dimension, illustrating the mechanism of Dynamic-Configurability.}
    \label{fig: Framework}
\end{figure*}

\subsection{Modeling of CogniSNN}

The RGA within CogniSNN is constructed using the WS and ER network generators designed by~\citet{2019_xie_exploring}, and is formalized as a Directed Acyclic Graph (DAG). Formally, the DAG is defined as $G = \{\mathcal{V}, E\}$, where $\mathcal{V}$ refers to the node set and $E$ is the edge set. As illustrated in the RGA-based SNN module in \textcolor{blue}{Fig.}~\ref{fig: Framework}, each node $v_i \in \mathcal{V}$ corresponds to a ResNode, while each edge $e \in E$ corresponds to a synaptic connection between nodes, with a learnable weight. The adjacency matrix $A$ of $G$ encodes the connectivity, where $A_{ij}$ = $w_{ij}$, if a synaptic connection exists from $v_j$ to $v_i$ with weight $w_{ij}$, and $A_{ij} = 0$ otherwise. For a given node $v_j$, let $\mathcal{N}_j$ represent the set of its predecessor nodes, with edges directed toward $v_j$. The input of $v_j$ at timestep $t$ is obtained as the weighted sum of its upstream nodes' output, expressed as:
\begin{align}
    I_{j}[t] = \sum_{i \in \mathcal{N}_j} \phi(A_{i,j}) \times AP(SP(O_{i}[t])).
    \label{Eq7: input}
\end{align}%
Here, $O_{i}[t]$ refers to the output of $v_i$ at timestep $t$. $\phi(\cdot)$ denotes the synapse-gating function, with candidates including sigmoid, softmax, tanh, etc. In this work, we adopt the sigmoid function. The $SP$ is Standard Pooling, which represents the inevitable downsampling operation in classification tasks. We define $SP$ as follows:
\begin{align}
    SP(O_{i}[t]) = \left\{\begin{array}{ll}
    AvgP(O_{i}[t], \kappa), & D(O_{i}[t]) \geq \eta \\
    O_{i}[t] & Otherwise
    \end{array}, \right.
    \label{Eq8:Standard pooling}
\end{align}%
\noindent where $D(\cdot)$ and $AvgP(\cdot)$ stand for spatial dimension and average pooling, respectively. $\eta$ specifies the minimum allowable spatial dimension within the RGA-based SNN (typically set to 1 in classification tasks), and $\kappa$ is the pooling kernel size (typically set to 2). $AP$ represents Adaptive Pooling, a mechanism introduced to resolve spatial dimension mismatches arising from random connectivity and downsampling. Further details will be provided in the subsequent section.

\subsubsection{ResNode}
We employ the ResNode to mitigate network degradation arising from excessively long neural pathways. As depicted in \textcolor{blue}{Fig.}~\ref{fig: Framework}, we define a fundamental building block, the $ConvBNSN$ triplet, which consists of a Convolutional (Conv) layer, a Batch Normalization (BN) layer, and a Spiking Neuron (SN) layer, i.e. $ConvBNSN(x) = SN(BN(Conv(x)))$. Each ResNode comprises two such triplets, denoted as $ConvBNSN_i^1$ and $ConvBNSN_i^2$ for a given node $v_i$. As outlined in \textcolor{blue}{Section}~\ref{sec:ResInSNN}, existing residual mechanisms in SNNs are predominantly based on the arithmetic $ADD$ operation, which results in unavoidable non-spike propagation. To overcome this limitation, we propose replacing the $ADD$ with a logical $OR$ gate.
 
The structure of the ResNode is illustrated in \textcolor{blue}{Fig.}~\ref{fig: Framework}. Each node aggregates outputs from its predecessor nodes and computes its input via weighted summation. Since this weighted aggregation converts discrete spike signals into real-valued data, the first triplet $ConvBNSN^1$ is utilized to transform these signals back into the spike domain.

To address network degradation, we introduce an advanced skip connection mechanism. Specifically, the output of the first triplet, $ConvBNSN^1$, is treated as the identity mapping (the skip connection), while the output of the second triplet, $ConvBNSN^2$, serves as the residual mapping (the processed features). These two components are combined using a logical $OR$ operation. The resulting binary signal serves as the node’s final output, which is then propagated to subsequent nodes. The entire process is formalized in Eq.~\ref{eq:skip}:
\begin{equation}
    \begin{split}
        O_{i}^{1}[t] &= ConvBNSN_{i}^{1}(I_{i}[t]) \\[10pt] 
        O_{i}^{2}[t] &= ConvBNSN_{i}^{2}(O_{i}^{1}[t]) \\[10pt] 
        O_{i}[t] &= OR(O_{i}^{2}[t], O_{i}^{1}[t]) \\
                 &= O_{i}^{2}[t] + O_{i}^{1}[t] - O_{i}^{2}[t] \odot O_{i}^{1}[t]
    \end{split}.
    \label{eq:skip}
\end{equation}
At timestep $t$, $I_{i}[t]$ is the input to node $v_{i}$. $O_{i}^{1}[t]$ and $O_{i}^{2}[t]$ refer to the outputs of the first triplet $ConvBNSN_{i}^{1}$ and the second triplet $ConvBNSN_{i}^{2}$, respectively. $OR(\cdot)$ represents our proposed pure spiking residual mechanism, termed the OR Gate, which employs logical OR operations. During backpropagation, the OR Gate is treated as a continuous function for training. Consequently, the final output $O_{i}[t]$ will emit a spike if either the residual or the identity branch contains a spike.

To rigorously validate the theoretical soundness of the ResNode for deep network training, we first demonstrate its ability to degenerate into an identity mapping within ResNode.
\begin{property}
    The OR Gate can implement ResNode-level identity mapping.
\end{property}

\begin{proof}
    For a given $v_i$, let $O_i^2[t]$ represent the learnable residual mapping. If we initialize the BN weights and biases in the second triplet to zero, or alternatively enforce a sufficiently large attenuation factor and firing threshold, the output of this block can be suppressed, i.e., $O_i^2[t] \equiv 0$. Under this condition, the output of $v_i$ becomes $O_{i}[t] = OR(O_{i}^{2}[t], O_{i}^{1}[t]) = OR(0, O_{i}^{1}[t]) = O_{i}^{1}[t]$. This relationship holds consistently across all nodes. Thus, the final output $O_i[t]$ aligns perfectly with the input branch $O_i^1[t]$, demonstrating that the OR Gate can faithfully realize an identity mapping under these specific conditions.
\end{proof}

Building upon this structural foundation, we further investigate the gradient propagation characteristics of the OR Gate to verify its efficacy in stabilizing the training process.

\begin{corollary}
    The OR Gate operation can alleviate gradient vanishing and explosion.
    \label{corollary1}
\end{corollary}

\begin{equation}
    \begin{split}
        \frac{\partial\, O_i[t]}{\partial\, O_i^1[t]} 
        &=  \frac{\partial \, OR(O_{i}^{2}[t], O_{i}^{1}[t])}{\partial\,O_i^1[t]} \\ 
        &= \frac{\partial \, (O_{i}^{2}[t] + O_{i}^{1}[t] - O_{i}^{2}[t] \odot O_{i}^{1}[t])}{\partial \,O_{i}^{1}[t]} \\
        &= \frac{\partial\, O_i^2[t]}{\partial \,O_{i}^{1}[t]} + 1 -O_i^2[t] - O_i^1[t]\cdot\frac{\partial\, O_i^2[t]}{\partial \,O_{i}^{1}[t]} \\
        &=(1-O_i^2[t])+(1-O_i^1[t])\cdot \frac{\partial\, O_i^2[t]}{\partial \,O_{i}^{1}[t]} \\   
        &=\left\{\begin{array}{ll}
1, & O_i^1[t] =1;\, O_i^2[t]=0 \\[6pt]
0, & O_i^1[t] =1;\, O_i^2[t]=1 \\[6pt]
1+ \frac{\partial \, O_i^2[t]}{\partial \,O_i^1[t]}, & O_i^1[t] =0;\, O_i^2[t]=0 \\[6pt]
\frac{\partial \, O_i^2[t]}{\partial \,O_i^1[t]},  & O_i^1[t] =0;\, O_i^2[t]=1 \\[6pt]
\end{array} \right.\\
    \end{split}.
\end{equation}
In ResNode $v_i$, at timestep $t$, we compute the derivative of the output $O_i[t]$ with respect to $O_i^1[t]$. The $OR$ gate yields four distinct gradient scenarios. When $O_i^1[t]=1$: if $O_i^2[t]=0$, the gradient is 1; if $O_i^2[t]=1$, the gradient is 0. Conversely, when $O_i^1[t]=0$: if $O_i^2[t]=0$, the gradient becomes $1+\frac{\partial\, O_i^2[t]}{\partial\,O_i^1[t]}$; if $O_i^2[t]=1$, the gradient simplifies to $\frac{\partial\, O_i^2[t]}{\partial\,O_i^1[t]}$. These gradients dynamically adjust based on spiking activity.

Specifically, in the absence of spike firing ($O_i^1[t]=0, O_i^2[t]=0$), the gradient term $1+\frac{\partial\, O_i^2[t]}{\partial\,O_i^1[t]}$ provides a strong error signal to drive weight updates and encourage firing. In contrast, when firing activity is excessive, the gradient diminishes or drops to zero, thereby stabilizing weights and suppressing redundant discharges.

It is worth noting that this specific gradient analysis focuses on the interaction within the ResNode. Since the input to $ConvBNSN^1$ originates from the weighted sum of multiple upstream nodes, we simplify the optimization landscape by focusing on the gradient flow through the OR Gate itself.

Finally, we address the fundamental signal integrity issues prevalent in conventional residual SNNs, specifically regarding value accumulation and the discrete nature of spikes.

\begin{property}
    The OR Gate operation resolves the issues of unbounded value accumulation and non-spike transmission.
    \label{property1}
\end{property}

In conventional $ADD$-based residual mechanisms, both in SNNs and ANNs, residuals are added to the identity mapping, leading to continuous value accumulation along the residual branch. This can result in the unbounded amplification of outputs. In contrast, the $OR$ gate performs a logical operation, ensuring that the output of each ResNode remains strictly binary (0 or 1), effectively preventing uncontrolled value growth. Furthermore, throughout identity branching, connection computation, and final node outputs, information is strictly preserved in the form of discrete spikes. This eliminates non-spike propagation and maintains the intrinsic, event-driven nature of SNNs.

\subsubsection{Adaptive Pooling Strategy}
In the proposed RGA, a node receiving inputs from multiple predecessor nodes may encounter feature maps with varying spatial dimensions. This discrepancy arises from the inconsistent application of Standard Pooling (SP) operations along different incoming pathways. To resolve this, we introduce an Adaptive Pooling (AP) mechanism applied prior to the integration of upstream information.

Specifically, for each node, the system first identifies the minimum spatial dimension among all incoming feature maps. Subsequently, all incoming feature maps are downsampled to align with this minimum dimension using average pooling. Let $v_j$ aggregate information from all its predecessor nodes $\mathcal{N}_j$. The input spatial dimension $D(I_j[t])$ is determined by:
\begin{align}
    D(I_j[t]) = \min \left \{ D(O_{i}[t])\mid i \in \mathcal{N}_j \right \}.
\end{align}
Consequently, the Adaptive Pooling (AP) is formally defined as:
\begin{align}
    AP(O_{i}[t]) = AvgP(O_{i}[t],\left\lfloor \frac{D(O_{i}[t])}{D(I_j[t])} \right\rfloor ),\, i\in\mathcal{N}_j.
    \label{eq: TP}
\end{align}%

This ensures that every output from predecessor node $v_i$ is resized to match the target spatial dimension $D(I_j[t])$ via average pooling with a dynamically calculated kernel size of $\left\lfloor \frac{D(O_{i}[t])}{D(I_j[t])} \right\rfloor$.

\subsection{Key Pathway-Based Learning without Forgetting}

To realize Pathway-Reusability in CogniSNN, we draw inspiration from the Learning without Forgetting (LwF)~\citep{2017_li_learning} paradigm in continual learning. Analogous to the biological brain, this approach mitigates catastrophic forgetting by regularizing internal parameters rather than replaying historical data. Our objective is to fine-tune specific neural pathways using this method, thereby evaluating their adaptability to diverse task scenarios. A pivotal challenge, however, lies in identifying the appropriate neural pathways within the graph $G$. As discussed in \textcolor{blue}{Section}~\ref{sec:bc}, betweenness centrality (BC) in graph theory quantifies the influence of nodes and edges on global information flow. Extending this concept, we define the Pathway Betweenness Centrality, $BC(p)$, as the cumulative sum of the BC values of its constituent nodes and edges.

\begin{definition}
    $BC(p) = \sum_{i=1}^{l_p+1} BC(v_{i}) + \sum_{j=1}^{l_p} BC(e_{j}).$
    \label{eq: C_B(p)}
\end{definition}

\noindent where  $p = \{ v_{1}, e_{1},v_{2},e_{2},...,e_{l_p},v_{l_p+1} \}$, denotes a specific neural pathway, $l_p$ refers to the length of $p$. According to Eqs.~\ref{Eq5.node BC} and \ref{Eq6.edge BC}, $BC(p)$, $BC(v)$, and $BC(e)$ represent the BC of neural pathway $p$, node $v$, and edge $e$, respectively.

Next, we sort all neural pathways $P$ in graph $G$ in descending order based on their BC:
\begin{align}
    P = \{ p_{1},p_{2},...,p_{|P|} \},
    \label{Eq.pk}
\end{align}%
\noindent where $BC(p_{1})\ge BC(p_{2}) ...\ge BC(p_{|P|})$, $|P|$ denotes the  total number of paths.

Having identified and ranked all neural pathways, we propose leveraging specific sub-structures within the RGA-based SNN to mirror the reusability of neural pathways in the brain, which reshapes functions through neural reorganization to accommodate novel experiences. Diverging from conventional continual learning and transfer learning techniques, our approach focuses on fine-tuning parameters within selective sub-structures, termed Key Pathways. These pathways correspond to neural connections essential for extracting generalizable features across diverse scenarios. Specifically, we define these key pathways $\mathcal{C}$ as follows:
\begin{definition}
    $\mathcal{C} \triangleq \{p_{k}| k \in [1, K], K \leq |P| \}$ if scenarios are similar. Otherwise, $\mathcal{C} \triangleq \{ p_{k} | k \in [|P|- K + 1, |P|], K \leq |P| \}$.
    \label{def: critical paths}
\end{definition}

The selection of key pathways is designed to preserve adaptability across different scenarios. For similar scenarios, the key pathway set consists of the top-$K$ neural pathways with the highest BC, as high-BC paths typically encode robust, shared features. Conversely, for scenarios with low similarity, the key pathway set corresponds to the bottom-$K$ neural pathways with the lowest BC, minimizing interference with established knowledge. In this work, to simplify the analysis, we set $K=1$. The complete procedure is outlined in Algorithm \ref{alg:cp-lwf}.

Formally, let $X^+$ and $Y^+$ denote the data and labels for the new task, respectively. Let $CogniSNN$ be the model pre-trained on the old task. $\mathcal{C}$ represents the selected key pathway, $\mathcal{G}$ denotes the entire graph, and $\mathcal{E}$ is the rounds of KP-LwF. $\theta$ represents the parameters, and $\hat{\theta}$ signifies parameters undergoing optimization. The algorithm's output is $CogniSNN^+$, a model capable of retaining knowledge from both old and new tasks.

The algorithm proceeds as follows: First (Line 1), we initialize $CogniSNN^+$ by cloning $CogniSNN$. Second (Line 2), we align the classification layer $f^+$ of $CogniSNN^+$ with the classification layer $f$ of the original model. Third (Line 3), we freeze all parameters in $CogniSNN^+$ except for those belonging to the key pathway $\mathcal{C}$ and the classifier $f^+$.

Subsequently, in each round of KP-LwF (Lines 6-8), we utilize the new data $x^+$ to generate three outputs: 1) $Y_o$ from the old model, serving as a soft label (distillation target) to retain historical information; 2) $Y_o'$ from the new model corresponding to the old task logits; 3) $Y_n'$ from the new model corresponding to the new task logits. We then compute the total loss function , comprising three terms: $L_{old}$, a distillation loss calculated using $Y_o$ and $Y_o'$ to prevent forgetting of old knowledge; $L_{new}$, a task loss calculated using  $Y^+$ and $Y_n'$ to ensure learning of the new task; and $R$, a regularization term controlling the magnitude of parameter updates. In this paper, we use the CrossEntropy loss function.

\begin{algorithm}[tb]
    \caption{The Key Pathway-based LwF algorithm}
    \label{alg:cp-lwf}
    \textbf{Input}: $X^+$, $Y^+$, $CogniSNN$, $\mathcal{C}$, $G$, $\mathcal{E}$\\
    \textbf{Output}: $CogniSNN^+$
    \begin{algorithmic}[1] 
        \STATE $CogniSNN \rightarrow CogniSNN^+$.
        \STATE Align the fully-connected layers $f$ and $f^+$.
        \STATE Freeze all parameters in $CogniSNN^+$ except for $\theta_{\mathcal{C}}$ and $\theta_{f^+}$.
        \FOR{epoch = 1 \textbf{to} $\mathcal{E}$}
            \FOR{$\forall (x^+, y^+)$ in $(X^+, Y^+)$}

            \STATE $Y_o = CogniSNN(x^+,\theta_G, \theta_{f})$.
            \STATE $Y_o^{'}= CogniSNN^+(x^+,\hat{\theta}_{G}, \hat{\theta}_{f})$.
            \STATE $Y_n^{'} = CogniSNN^+(x^+,\hat{\theta}_{G}, \hat{\theta}_{f^+})$.
            \STATE $loss = \lambda \cdot L_{old}(Y_o, Y_o^{'}) + L_{new}(y^+, Y_n^{'}) + R(\hat{\theta}_G, \hat{\theta}_f , \hat{\theta}_{f^+}).$
            \STATE Update $\hat{\theta}_{\mathcal{C}}$ and $
            \hat{\theta}_{f^+}$ using STBP.
            \ENDFOR
        \ENDFOR
        \STATE \textbf{return} $CogniSNN^+$
    \end{algorithmic}
\end{algorithm}

\begin{algorithm}[tb]
    \caption{The Dynamic Growth Learning algorithm}
    \label{alg:dgl}
    \textbf{Input}: $X$, $Y$, $CogniSNN$, $P$, $\mathcal{E}$, $T$\\
    \textbf{Output}: Trained $CogniSNN$
    \begin{algorithmic}[1]
        \FOR{$epoch = 1$ \textbf{to} $\mathcal{E}$}
            \FOR{\textbf{each} $(x, y)$ \textbf{in} $(X, Y)$}
                \STATE Initialize spike accumulator $O_{sum} = 0$.
                \FOR{$t = 1$ \textbf{to} $T$}
                    \STATE Calculate active subgraph $\mathcal{P}^{(t)}$ by Eq.~\ref{eq:subgraph}.
                    \STATE $o_t = CogniSNN(\mathcal{P}^{(t)}, x[t])$.
                    \STATE $O_{sum} = O_{sum} + o_t$.
                \ENDFOR 
                \STATE $O_{mean} = O_{sum} / T$.
                \STATE Calculate loss $\mathcal{L}$ using $O_{mean}$ and label $y$.
                \STATE Update parameters of $CogniSNN$ using STBP.
            \ENDFOR
        \ENDFOR
        \STATE \textbf{return} $CogniSNN$
    \end{algorithmic}
\end{algorithm}

\subsection{Dynamic Growth Learning Algorithm}

To further enhance the robustness of CogniSNN, we simulate the dynamic growth mechanism of neural pathways observed in the biological brain. This evolutionary process is visualized in the bottom part of \textcolor{blue}{Fig.}~\ref{fig: Framework}. Specifically, during the training phase, the network operates on a progressively expanding set of neural pathways, denoted as $\mathcal{P}^{(t)}$ at timestep $t$.  This dynamic set is defined as:
\begin{align}
    \mathcal{P}^{(t)} &= \left \{ p_k \,|\, 1 \le k \le q(t)  \right \} , \, \\ q(t)&=\begin{cases}
 t\cdot \left \lfloor \frac{|P|}{T}  \right \rfloor , & 1\le t < T \\
 |P|, & t=T
 \label{eq:subgraph}
\end{cases},
\end{align}

\noindent where the nested relationship satisfies $\mathcal{P}^{(1)} \subset \mathcal{P}^{(2)} \subset \cdot\cdot\cdot \subset \mathcal{P}^{(T)}$. Similar to Eq.~\ref{Eq.pk}, $p_k$ is sorted from high to low based on BC.

Here, $q(t)$ denotes the cumulative number of active neural pathways at time $t$. When $t = T$, the set encompasses all pathways, i.e., $\mathcal{P}^{(t)} =  \left \{ p_k \,|\, 1 \le k \le |P| \right \}  = P$. 

Existing Spiking Neural Networks are typically trained on a static network architecture that remains fixed across all $T$ simulation timesteps. In contrast, our proposed DGL algorithm trains CogniSNN on a time-varying active subgraph $\mathcal{P}^{(t)}$ at each timestep $t$, thereby simulating the progressive expansion of biological neural networks. During inference, all neural pathways are utilized. Among these, pathways not fully trained under complete timesteps significantly enhance the network's robustness and deployment flexibility. It is worth noting that we do not explicitly simulate the apoptosis of neural pathways; during the learning process, the synaptic weights of irrelevant or redundant neurons naturally converge toward near-zero values, which effectively functions as an implicit pathway apoptosis mechanism.

The implementation details are outlined in Algorithm \ref{alg:dgl}. Unlike conventional SNN training, at each timestep $t$, we first determine the graph topology corresponding to $\mathcal{P}^{(t)}$. This active subgraph is then utilized to govern the forward propagation of information.

\section{Experiment}

\subsection{Experiment Setting}

To ensure statistical reliability and reproducibility, we fix the random seed controlling the graph generation to stabilize the network topology. Furthermore, we conduct five independent trials using different random seeds to initialize network weights, reporting the mean and standard deviation. All experiments are implemented using PyTorch and the SpikingJelly framework on an NVIDIA RTX 4090 GPU. To comprehensively validate the advantages of CogniSNN in terms of \textbf{Neuron-Expandability}, \textbf{Pathway-Reusability}, and \textbf{Dynamic-Configurability}, we design four categories of experiments: classification on neuromorphic and static datasets; continual learning under diverse scenarios; anti-interference robustness evaluation; and deployment flexibility assessment. Additionally, we conduct extensive ablation studies and energy consumption analyses to provide a thorough and fair comparison.

\textbf{1) Classification on Neuromorphic and Static Datasets:} In this task, we evaluate our model on three neuromorphic datasets, DVS-Gesture~\citep{2017_amir_low}, CIFAR10-DVS~\citep{2017_li_cifar10}, and N-Caltech101~\citep{2015_orchard_converting}, and one static dataset, Tiny-ImageNet~\citep{2015_le_tiny}. 
\begin{itemize}
    \item \textbf{DVS-Gesture:} This dataset targets gesture recognition, capturing 11 gesture categories from different individuals. It contains 1,464 samples (1,176 for training and 288 for testing), with a spatial resolution of $128 \times 128$. We process the dataset directly using the tools provided by SpikingJelly without additional preprocessing or data augmentation.
    \item \textbf{CIFAR10-DVS:} As the neuromorphic version of the static CIFAR10 dataset, it contains the same 10 categories with 1,000 event-stream samples per category. The original resolution of $128 \times 128$ is downsampled to $64 \times 64$. Following standard protocols, we split the dataset into training and testing sets with a 9:1 ratio. Data augmentation techniques, identical to those in other baselines, are applied during training.
    \item \textbf{N-Caltech101}: This dataset is the neuromorphic counterpart of Caltech101, comprising 101 categories and 8,709 samples with a spatial resolution of $180 \times 240$. For consistency, we resize all samples to $128 \times 128$ using bilinear interpolation without extra data augmentation. The dataset was split into training and testing sets in a 9:1 ratio.
    \item \textbf{Tiny-ImageNet}: This is a reduced version of ImageNet-1000, containing 200 categories with 600 images per category. The dataset is divided into training, validation, and test sets with a 10:1:1 ratio. All images are resized to $64 \times 64$, reducing computational costs and facilitating performance evaluation in resource-constrained environments.
\end{itemize}

We generate two types of graph-structured networks driven by WS and ER paradigms, respectively, each containing 7 ResNodes. The models are trained with a batch size of 8 using Stochastic Gradient Descent (SGD) with a momentum of 0.9 and no weight decay. For the DVS-Gesture and CIFAR10-DVS datasets, we employ a Cosine Annealing Learning Rate Scheduler with an initial learning rate of 0.001 and a maximum period of $T_{\text{max}} = 64$, training for 300 epochs. For the other datasets, models are trained for 192 epochs with an initial learning rate of 0.001, which decayed by a factor of 10 every 64 epochs.

\textbf{2) Continual Learning under Different Scenarios:} Following the experimental protocol described in the LwF framework, we first pre-train the network on CIFAR100 and subsequently perform class-incremental learning on either CIFAR10 or MNIST. We employ the Fréchet Inception Distance (FID)~\citep{2017_heusel_gans}, a metric widely used in image generation, to quantify task similarity, establishing a threshold of 50. The FID value between CIFAR100 and MNIST is 2341.2, far surpassing this limit and signifying distinct tasks. Conversely, the FID between CIFAR100 and CIFAR10 stands at 24.5, classifying them as similar tasks. Therefore, we treat the transition from CIFAR100 to MNIST as a dissimilar scenario, while CIFAR100 to CIFAR10 represents a similar scenario. 

During the KP-LwF process, we ensure that the network exhibits a comparable level of forgetting on the old task (CIFAR100), thereby isolating and assessing its ability to learn new tasks. To guarantee sufficiently rich neural pathways for sampling, we employ ER- and WS-driven random graph networks with 32 ResNodes for these experiments.

\textbf{3) Anti-Interference Evaluation:}  We further evaluate the robustness conferred by the dynamic growth training algorithm using the DVS-Gesture dataset. Specifically, the network is trained on clean neuromorphic data and tested on perturbed sets. We examine two types of robustness: random noise robustness (including Salt-and-Pepper noise and Poisson noise) and frame-loss robustness. For Salt-and-Pepper noise, the intensity is controlled by $\rho$, where the proportion of the noisy region is calculated as $\rho \times 0.02$. For Poisson noise, the interference strength $\rho$ denotes the noise scaling factor. For frame-loss robustness, the probability of randomly dropping frames is defined as $\rho \times 0.05$.

\textbf{4) Deployment Flexibility Assessment:} Directly trained SNNs typically suffer from the constraint of fixed timesteps, requiring the inference timestep to be consistent with the training timestep. Performance degrades significantly when inference steps are insufficient, severely limiting deployment on time-flexible neuromorphic chips. To evaluate adaptability, we train CogniSNN (with and without DGL algorithm) on DVS-Gesture and CIFAR10-DVS using 8 timesteps, and subsequently perform inference using fewer timesteps to assess performance retention.

\subsection{Neuron-Expandability}

\begin{table*}[ht]
\caption{Performance comparison with state-of-the-art methods on DVS-Gesture, CIFAR10-DVS, N-Caltech101, and Tiny-ImageNet datasets. We report results for both ER-driven and WS-driven variants of CogniSNN. ER-RGA-7 represents the ER-driven random graph architecture with 7 ResNodes, and WS-RGA-7 follows the same principle. The \textbf{bold} and \underline{underline} markers represent the best and second-best values. These baselines include  SEW-ResNet~\citep{2021_fang_deep}, 
ESTSformer~\citep{2025_lu_estsformer}, 
SAFormer~\citep{2025_zhang_combining},
STSA~\citep{2023_wang_spatial}, 
CML~\citep{2023_zhou_enhancing}, Spikformer~\citep{2022_zhou_spikformer}, SSNN~\citep{2024_ding_shrinking}, MLF~\citep{2022_feng_multi}, TT-SNN~\citep{2024_lee_tt}, NDA~\citep{2022_li_neuromorphic}, TIM~\citep{2024_shen_tim}, Joint SNN~\citep{2023_guo_joint}, SNASNet-Bw~\citep{2022_kim_neural}, and Online LTL~\citep{2022_yang_training}.} 
    \centering
    \renewcommand{\arraystretch}{1.2} 
    \setlength{\tabcolsep}{0pt} 
    \begin{tabular*}{\textwidth}{@{\extracolsep{\fill}}>{\centering\arraybackslash}p{0.14\textwidth}
                                                    >{\centering\arraybackslash}p{0.17\textwidth}
                                                    >{\centering\arraybackslash}p{0.20\textwidth}
                                                    >{\centering\arraybackslash}p{0.09\textwidth}
                                                    >{\centering\arraybackslash}p{0.08\textwidth}
                                                    >{\centering\arraybackslash}p{0.28\textwidth}}
        \toprule
        Dataset & Method & Network & Param(M) & T & Accuracy($\%$) \\
        \midrule
        \multirow{8}{*}{\makecell{DVS-Gesture}}
                           & SEW-ResNet & SEW-ResNet & 0.13 & 16 & 97.90 \\
                           & ESTSformer& EST-Sformer-2-256 & - & 16 & 98.26 \\
                           & SAFormer& SAFormer-2-256 & - & 16 & 98.3 \\
                           
                           & STSA & STS-Transformer-1-256& - & 16 & 98.38\\
                           & CML & Spikformer-4-384 & 2.57 & 16 & \underline{98.60} \\
                           & Spikformer & Spikformer-2-256 & 2.57 & 5 / 16 & 79.52 / 98.30 \\
                           & SSNN& VGG-9 & - & 5 / 8 & \underline{90.74} / \underline{94.91} \\

         \cmidrule{2-6}
                           & \multirow{2}{*}{CogniSNN(Ours)} & ER-RGA-7 &\multirow{2}{*}{0.13} & \multirow{2}{*}{5 / 8 / 16} & \textbf{94.78}$ \hspace{0.1em} \scriptstyle \pm \hspace{0.1em} 0.12$/ \textbf{95.81}$ \hspace{0.1em} \scriptstyle \pm \hspace{0.1em} 0.15$/
                            \textbf{98.61}$ \hspace{0.1em} \scriptstyle \pm \hspace{0.1em} 0.11$\\

                             &  &WS-RGA-7& &    &  \textbf{95.41}$ \hspace{0.1em} \scriptstyle \pm \hspace{0.1em} 0.21$/ \textbf{96.51}$ \hspace{0.1em} \scriptstyle \pm \hspace{0.1em} 0.09$/ 97.23$ \hspace{0.1em} \scriptstyle \pm \hspace{0.1em} 0.34$\\

        \cmidrule{1-6}
        
        \multirow{8}{*}{\makecell{CIFAR10-DVS}}
                            & MLF& VGG-9 & - & 5 & 67.07 \\
                            & SEW-ResNet & SEW-ResNet & 1.19 & 8 & 70.20 \\
                            & CML & Spikformer-4-384 & - & 10 & 80.5 \\
                           & SSNN& VGG-9 & - & 5 / 8 & 73.63 / 78.57 \\
                            & Spikformer & Spikformer-2-256 & 2.57 & 5 / 10 & 68.55 / 78.90\\
                           & STSA & STS-Transformer-1-256& - & 10 / 16 & \underline{78.28} / \underline{79.60}\\
        \cmidrule{2-6}
                           &\multirow{2}{*}{CogniSNN(Ours)} & ER-RGA-7 & \multirow{2}{*}{1.51} & \multirow{2}{*}{5 / 8} & \textbf{79.80}$ \hspace{0.1em} \scriptstyle \pm \hspace{0.1em} 0.10$ / \textbf{80.50}$ \hspace{0.1em} \scriptstyle \pm \hspace{0.1em} 0.20$ \\

                            &  & WS-RGA-7 & & & \textbf{79.00}$ \hspace{0.1em} \scriptstyle \pm \hspace{0.1em} 0.21$ /  \textbf{80.52}$ \hspace{0.1em} \scriptstyle \pm \hspace{0.1em} 0.13$\\
        \cmidrule{1-6}
      
        \multirow{7}{*}{\makecell{N-Caltech101}}
                           & Spikformer& Spikformer & 2.57 & 5 & 72.8 \\
                           & TT-SNN & ResNet34 & - & 6 & 77.80 \\
                           & NDA & VGG11 & - & 10 & 78.2 \\
                           & TIM  &Spikformer & - & 10 & 79.0 \\
                           & SSNN & VGG-9 & - & 5 / 8 & \underline{77.97}/ \underline{79.25} \\
        \cmidrule{2-6}
                           &\multirow{2}{*}{CogniSNN(Ours)}& ER-RGA-7& \multirow{2}{*}{1.51}& \multirow{2}{*}{5 / 8}& \textbf{80.64}$ \hspace{0.1em} \scriptstyle \pm \hspace{0.1em} 0.15$/ \textbf{79.32}$ \hspace{0.1em} \scriptstyle \pm \hspace{0.1em} 0.15$\\

                           & &WS-RGA-7&  &   & \textbf{78.62}$ \hspace{0.1em} \scriptstyle \pm \hspace{0.1em} 0.31$/ \textbf{80.23}$ \hspace{0.1em} \scriptstyle \pm \hspace{0.1em} 0.11$\\

        \cmidrule{1-6}
      
        \multirow{5}{*}{\makecell{Tiny-ImageNet}}
                           & Joint-SNN & VGG-16 & - & 4 & \underline{55.39} \\
                           & SNASNet-Bw & Searched & - & 5 & 54.60 \\
                           & Online LTL & VGG-13 & - & 6 & 55.37 \\
        \cmidrule{2-6}
                           & \multirow{2}{*}{CogniSNN(Ours)}& ER-RGA-7 & \multirow{2}{*}{1.52} & \multirow{2}{*}{4} & \textbf{55.41}$ \hspace{0.1em} \scriptstyle \pm \hspace{0.1em} 0.17$\\
                           & & WS-RGA-7 & &  & \textbf{55.82}$ \hspace{0.1em} \scriptstyle \pm \hspace{0.1em} 0.10$\\
        \bottomrule
    \end{tabular*}
    \label{tab:totalTable}
\end{table*}
We employ the OR Gate mechanism in conjunction with Adaptive Pooling to mitigate network degradation and dimensional mismatch issues arising from excessively deep neural pathways.  \textcolor{blue}{Table} \ref{tab:totalTable} presents a comprehensive performance comparison between CogniSNN and other state-of-the-art models. 

As observed on the DVS-Gesture dataset, our model outperforms both conventional CNN-based SNNs and the ADD-based SEW-ResNet. Notably, under the low-latency setting ($T=5$), the ER- and WS-driven CogniSNN variants achieve substantial accuracy gains of $4.0\%$ and $4.7\%$, respectively, compared to the SSNN baseline. At $T=8$, they continue to demonstrate improvements of $0.9\%$ and $1.6\%$, respectively. Under $T=16$, the ER-driven CogniSNN achieves an accuracy of $98.61\%$, performing comparably to the transformer-based CML.

On the CIFAR10-DVS dataset, under the short timestep constraint $T=5$, the ER- and WS-driven CogniSNN models secure remarkable accuracy improvements of $6.2\%$ and $5.4\%$, respectively, compared to the SSNN. Furthermore, at $T=8$, our model remains highly competitive; specifically, the WS-driven variant achieves $80.52\%$, which is comparable to the state-of-the-art CML model.

On N-Caltech101, CogniSNN consistently delivers comprehensive improvements. Specifically, at $T=5$, the ER-driven and WS-driven variants achieve accuracy gains of $2.6\%$ and $0.7\%$, respectively. At $T=8$, the WS-driven model reaches $80.23\%$, outperforming state-of-the-art methods such as TIM ($79.0\%$) and NDA ($78.2\%$). Similarly, on the static Tiny-ImageNet dataset, CogniSNN demonstrates highly competitive performance, with the WS-driven variant achieving the highest accuracy of $55.82\%$ at $T=4$.

It is worth noting that while Transformer-based SNNs often struggle in low-latency regimes due to their reliance on long-term temporal dependencies, CogniSNN exhibits robust performance even at short timesteps. This underscores its superior capability in extracting spatial features from neuromorphic data, a finding further corroborated by the visualizations presented in the subsequent section.

\subsection{Pathway-Reusability}

\begin{figure}
    \centering
    \includegraphics[width=0.49\textwidth]{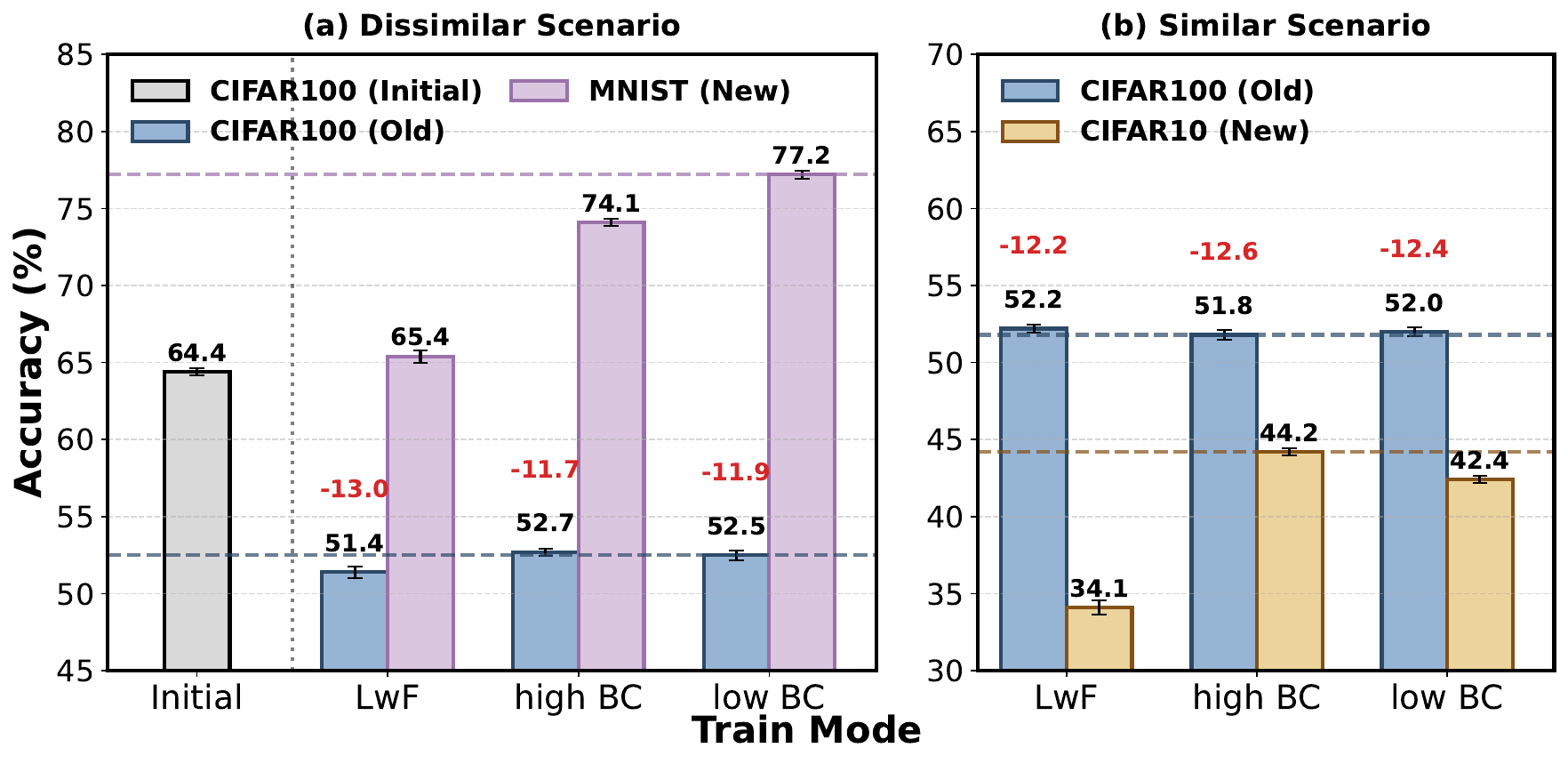}
    \caption{Performance comparison of the ER-driven CogniSNN across different continual learning scenarios. (a) Dissimilar Scenario (CIFAR100 $\rightarrow$ MNIST) and (b) Similar Scenario (CIFAR100 $\rightarrow$ CIFAR10). The red values indicate the accuracy drop on the old task  compared to the initial baseline.}
    \label{fig:CL-ER}
\end{figure}

\begin{figure}
    \centering
    \includegraphics[width=0.49\textwidth]{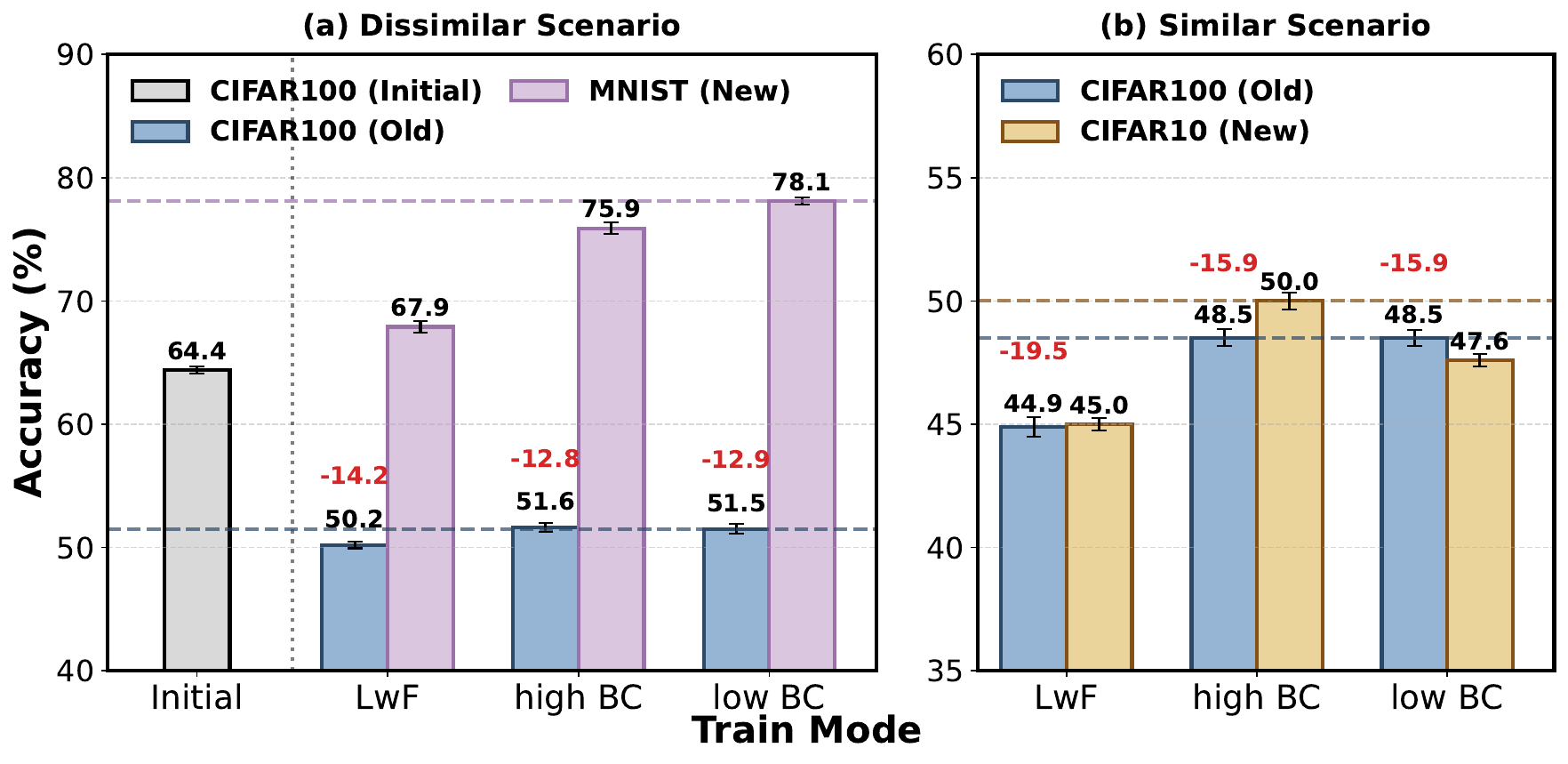}
     \caption{Performance comparison of the WS-driven CogniSNN across different continual learning scenarios. (a) Dissimilar Scenario (CIFAR100 $\rightarrow$ MNIST) and (b) Similar Scenario (CIFAR100 $\rightarrow$ CIFAR10). The red values indicate the accuracy drop on the old task  compared to the initial baseline.}
    \label{fig:CL-WS}
\end{figure}

We apply the KP-LwF algorithm to both WS- and ER-driven CogniSNN models, each configured with 32 nodes, across two distinct scenarios. In this context, Vanilla LwF refers to the standard LwF algorithm where the entire RGA-based SNN is fine-tuned, as opposed to selectively updating specific neural pathways. As shown in \textcolor{blue}{Figs.}~\ref{fig:CL-ER} and \ref{fig:CL-WS}, both ER- and WS-driven CogniSNN models exhibit consistent patterns.

For the ER-driven CogniSNN, when adapting to dissimilar scenarios (CIFAR100 to MNIST), learning new knowledge via low-BC neural pathways achieves a $3.1\%$ advantage over high-BC neural pathways under comparable levels of forgetting. In contrast, when adapting to similar scenarios (CIFAR100 to CIFAR10), fine-tuning high-BC neural pathways via KP-LwF outperforms fine-tuning low-BC pathways by $1.8\%$ in terms of new knowledge acquisition. Importantly, regardless of which neural pathway is selected for training, KP-LwF demonstrates a performance advantage of nearly $10.0\%$ over the Vanilla LwF baseline. This suggests that the benefits derived from diverse pathways are not merely a function of the parameter scale within different pathways, but rather stem from their topological properties. For WS-driven CogniSNN, the same phenomenon is observed.

These results lead to two key conclusions. First, in CogniSNN, high-BC neural pathways play a dominant role in encoding essential and transferable features from the initial task. Thus, when facing similar scenarios, reusing and fine-tuning these pathways leads to superior performance. Second, low-BC neural pathways, which are less central and primarily encode marginal features, prove more effective when adapting to dissimilar scenarios, where flexibility and reduced dependency on prior task features are advantageous. This behavior mirrors the human brain’s strategy of leveraging strong pathways for familiar tasks while recruiting less active pathways to accommodate novel ones, thereby providing experimental validation of our hypothesis.

\subsection{Dynamic-Configurability}

As detailed in \textcolor{blue}{Table} \ref{tab:robustness}, the proposed DGL algorithm demonstrates superior robustness against various types and levels of random noise interference. Here, the Baseline refers to a standard CogniSNN trained on the static, full random graph across all timesteps. Observations indicate that, regardless of the noise type—Salt-and-Pepper (SP) or Poisson (PN)—the model trained with dynamic growth consistently exhibits greater robustness compared to the baseline, particularly as interference intensity escalates.  Specifically, under SP perturbation at an intensity level of 8, the performance advantage widens to $16.38\%$. Similarly, under PN perturbation at intensity 8, the advantage reaches approximately $9.79\%$. In frame-loss scenarios, with an intensity of 8, the model outperforms the baseline by a margin of $10.10\%$.

\begin{figure}
	\centering
		\includegraphics[width=0.48\textwidth]{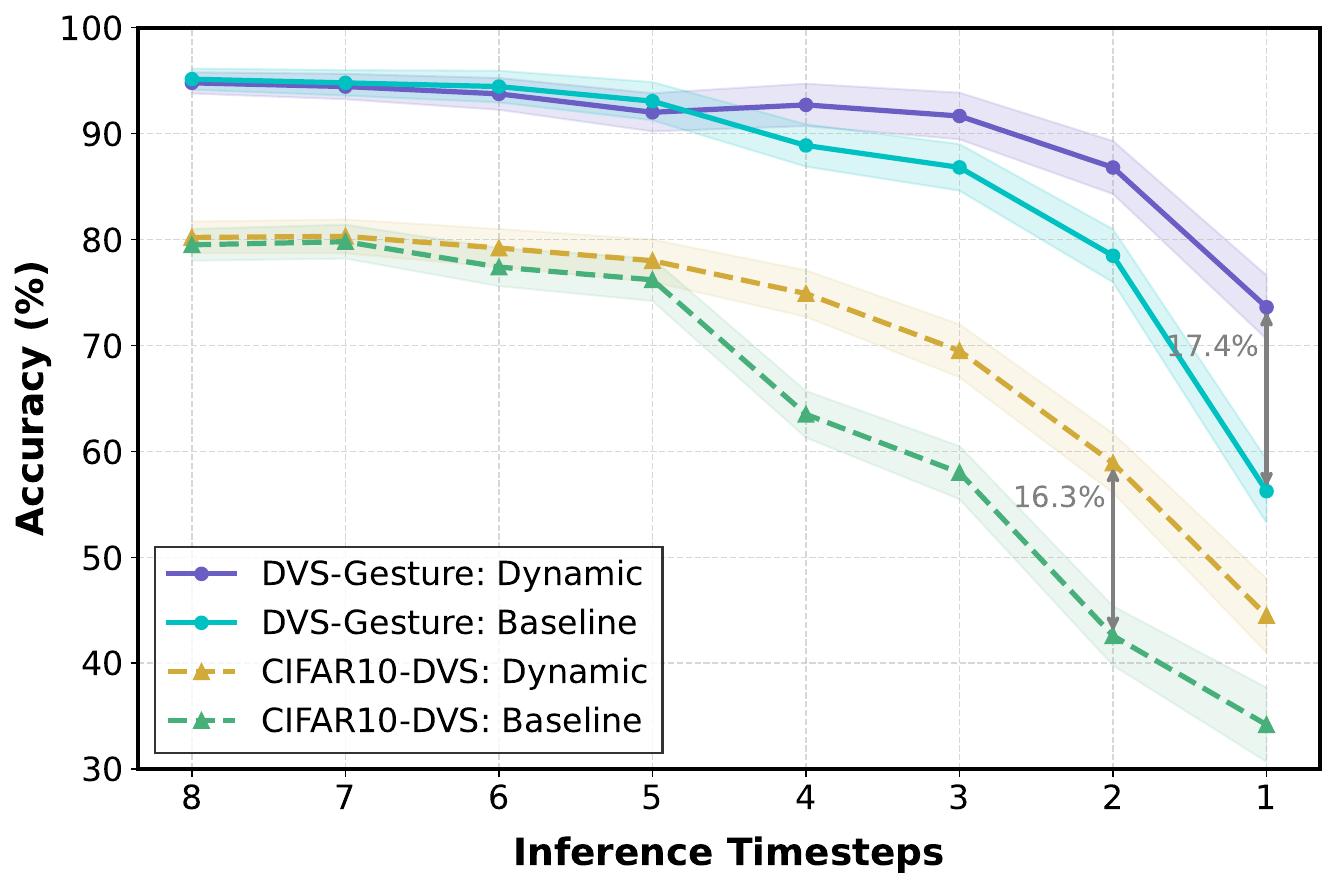}
	\caption{Accuracy comparison under reduced inference timesteps on DVS-Gesture and CIFAR10-DVS. ``Dynamic'' denotes the CogniSNN trained via the proposed DGL algorithm, while ``Baseline'' represents the static counterpart.}
	\label{fig: fixed-timestep}
\end{figure}

Regarding deployment flexibility, we compare the model trained via the DGL algorithm against the baseline trained via standard methods where train timestep is 8. The results in \textcolor{blue}{Fig.}~\ref{fig: fixed-timestep} clearly illustrate that when the inference timestep is reduced to less than half of the training duration, the baseline model's performance suffers a precipitous decline. In contrast, the dynamic model exhibits a much more graceful degradation. Remarkably, on the DVS-Gesture dataset, we achieve a $17.4\%$ performance retention over the baseline when performing inference with a single timestep. Similarly, on CIFAR10-DVS, we secure a $16.3\%$ advantage at two timesteps. These advantages stem from the fact that certain subgraphs during DGL training are adapted to infer effectively under temporally constrained conditions. In contrast, subgraphs trained under traditional paradigms have never undergone similar experiences.


\begin{table*}[ht]
\centering
\caption{Robustness comparison on the DVS-Gesture dataset under varying perturbation intensities ($\rho$). The evaluation includes three interference types: Salt-and-Pepper noise (SP), Poisson Noise (PN), and Frame Loss (FL). ``Dynamic'' denotes the CogniSNN trained via the proposed DGL algorithm, while ``Baseline'' represents the static counterpart. \colorbox{gray!25}{\textbf{Bold and Grey values}} indicate instances where the DGL algorithm outperforms the baseline.}
\label{tab:robustness}

\setlength{\tabcolsep}{1.5pt}

\resizebox{\textwidth}{!}{
\begin{tabular}{lllcccccccccc}
\toprule
\multirow{2}{*}{\textbf{Dataset}} & \multirow{2}{*}{\textbf{Noise}} & \multirow{2}{*}{\textbf{Model}} & \multicolumn{10}{c}{\textbf{Perturbation Intensity ($\rho$)}} \\
\cmidrule(lr){4-13}
 &  &  & \textbf{0} & \textbf{1} & \textbf{2} & \textbf{3} & \textbf{4} & \textbf{5} & \textbf{6} & \textbf{7} & \textbf{8} & \textbf{9} \\
\midrule

\multirow{6}{*}{\rotatebox{90}{DVS-Gesture}} 
& \multirow{2}{*}{SP} 
  & Baseline & 96.18$ \hspace{0.1em} \scriptstyle \pm \hspace{0.1em} 0.00$ & 94.72$ \hspace{0.1em} \scriptstyle \pm \hspace{0.1em} 1.13$ &93.96$ \hspace{0.1em} \scriptstyle \pm \hspace{0.1em} 1.00$ & 90.14$ \hspace{0.1em} \scriptstyle \pm \hspace{0.1em} 1.06$ & 83.33$ \hspace{0.1em} \scriptstyle \pm \hspace{0.1em} 1.01$ & 72.01$ \hspace{0.1em} \scriptstyle \pm \hspace{0.1em} 1.66$ & 56.81$ \hspace{0.1em} \scriptstyle \pm \hspace{0.1em} 1.64$ & 38.26$ \hspace{0.1em} \scriptstyle \pm \hspace{0.1em} 2.30$ & 21.81$ \hspace{0.1em} \scriptstyle \pm \hspace{0.1em} 0.86$ & 14.58$ \hspace{0.1em} \scriptstyle \pm \hspace{0.1em} 0.98$ \\
& & Dynamic  & 95.98$ \hspace{0.1em} \scriptstyle \pm \hspace{0.1em} 0.00$ & \greytextbf{94.93}$ \hspace{0.1em} \scriptstyle \pm \hspace{0.1em} 0.57$ & 93.52$ \hspace{0.1em} \scriptstyle \pm \hspace{0.1em} 1.01$ & \greytextbf{90.21}$ \hspace{0.1em} \scriptstyle \pm \hspace{0.1em} 1.20$ & \greytextbf{85.70}$ \hspace{0.1em} \scriptstyle \pm \hspace{0.1em} 1.42$ & \greytextbf{77.36}$ \hspace{0.1em} \scriptstyle \pm \hspace{0.1em} 1.79$ & \greytextbf{61.94}$ \hspace{0.1em} \scriptstyle \pm \hspace{0.1em} 1.56$ & \greytextbf{47.92}$ \hspace{0.1em} \scriptstyle \pm \hspace{0.1em} 0.95$ & \greytextbf{38.19}$ \hspace{0.1em} \scriptstyle \pm \hspace{0.1em} 0.81$ & \greytextbf{30.00}$ \hspace{0.1em} \scriptstyle \pm \hspace{0.1em} 1.48$ \\
\cmidrule{2-13}

& \multirow{2}{*}{PN} 
  & Baseline & 96.18$ \hspace{0.1em} \scriptstyle \pm \hspace{0.1em} 0.00$ &93.20$ \hspace{0.1em} \scriptstyle \pm \hspace{0.1em} 0.31$ & 86.25$ \hspace{0.1em} \scriptstyle \pm \hspace{0.1em} 1.36$ & 79.03$ \hspace{0.1em} \scriptstyle \pm \hspace{0.1em} 0.80$ & 68.40$ \hspace{0.1em} \scriptstyle \pm \hspace{0.1em} 1.72$ & 57.85$ \hspace{0.1em} \scriptstyle \pm \hspace{0.1em} 0.31$ & 50.90$ \hspace{0.1em} \scriptstyle \pm \hspace{0.1em} 1.64$ &43.33$ \hspace{0.1em} \scriptstyle \pm \hspace{0.1em} 1.96$ & 37.57$ \hspace{0.1em} \scriptstyle \pm \hspace{0.1em} 0.71$ & 33.82$ \hspace{0.1em} \scriptstyle \pm \hspace{0.1em} 1.34$ \\
& & Dynamic  & 95.98$ \hspace{0.1em} \scriptstyle \pm \hspace{0.1em} 0.00$ & 92.93$ \hspace{0.1em} \scriptstyle \pm \hspace{0.1em} 0.85$ & \greytextbf{88.12}$ \hspace{0.1em} \scriptstyle \pm \hspace{0.1em} 0.90$ & \greytextbf{81.81}$ \hspace{0.1em} \scriptstyle \pm \hspace{0.1em} 0.94$ & \greytextbf{73.96}$ \hspace{0.1em} \scriptstyle \pm \hspace{0.1em} 1.32$ & \greytextbf{66.67}$ \hspace{0.1em} \scriptstyle \pm \hspace{0.1em} 1.39$ & \greytextbf{58.68}$ \hspace{0.1em} \scriptstyle \pm \hspace{0.1em} 1.28$ & \greytextbf{52.08}$ \hspace{0.1em} \scriptstyle \pm \hspace{0.1em} 0.85$ & \greytextbf{47.36}$ \hspace{0.1em} \scriptstyle \pm \hspace{0.1em} 0.58$ & \greytextbf{42.08}$ \hspace{0.1em} \scriptstyle \pm \hspace{0.1em} 1.65$ \\
\cmidrule{2-13}

& \multirow{2}{*}{FL} 
  & Baseline  & 96.18$ \hspace{0.1em} \scriptstyle \pm \hspace{0.1em} 0.00$ & 95.83$ \hspace{0.1em} \scriptstyle \pm \hspace{0.1em} 0.85$ & 91.66$ \hspace{0.1em} \scriptstyle \pm \hspace{0.1em} 1.15$ & 83.12$ \hspace{0.1em} \scriptstyle \pm \hspace{0.1em} 1.24$ & 70.77$ \hspace{0.1em} \scriptstyle \pm \hspace{0.1em} 2.49$ & 60.00$ \hspace{0.1em} \scriptstyle \pm \hspace{0.1em} 0.79$ & 49.93$ \hspace{0.1em} \scriptstyle \pm \hspace{0.1em} 1.60$ & 39.72$ \hspace{0.1em} \scriptstyle \pm \hspace{0.1em} 0.76$ & 29.72$ \hspace{0.1em} \scriptstyle \pm \hspace{0.1em} 1.03$ & 22.64$ \hspace{0.1em} \scriptstyle \pm \hspace{0.1em} 0.45$ \\
& & Dynamic  & 95.98 $ \hspace{0.1em} \scriptstyle \pm \hspace{0.1em} 0.00$ & 95.41$ \hspace{0.1em} \scriptstyle \pm \hspace{0.1em} 0.80$ & 91.58$ \hspace{0.1em} \scriptstyle \pm \hspace{0.1em} 1.04$ & \greytextbf{84.93}$ \hspace{0.1em} \scriptstyle \pm \hspace{0.1em} 1.00$ & \greytextbf{74.58}$ \hspace{0.1em} \scriptstyle \pm \hspace{0.1em} 1.52$ & \greytextbf{64.93}$ \hspace{0.1em} \scriptstyle \pm \hspace{0.1em} 1.39$ & \greytextbf{54.51}$ \hspace{0.1em} \scriptstyle \pm \hspace{0.1em} 1.43$ & \greytextbf{45.35}$ \hspace{0.1em} \scriptstyle \pm \hspace{0.1em} 1.53$ & \greytextbf{37.98}$ \hspace{0.1em} \scriptstyle \pm \hspace{0.1em} 1.38$ & \greytextbf{32.74}$ \hspace{0.1em} \scriptstyle \pm \hspace{0.1em} 0.45$ \\

\bottomrule
\end{tabular}
}
\end{table*}

\subsection{Ablation Study}

To elucidate the specific sources of performance gains in classification tasks, we conduct two ablation studies: one examining the contribution of the random graph topology and the other analyzing the efficacy of the proposed OR Gate.

\textbf{1) Impact of Random Graph Structures:}
First, we quantify the extent to which random graph structures contribute to performance improvements compared to a traditional architecture. For this purpose, we establish a baseline model utilizing a conventional chain-like hierarchical architecture with seven ResNodes. We compare the performance of WS-driven CogniSNN, ER-driven CogniSNN, and this baseline across three datasets at $T=5$. As illustrated in \textcolor{blue}{Table}~\ref{tab:rga}, both WS- and ER-driven CogniSNN models consistently outperform the baseline by a margin exceeding $2\%$. This advantage is likely attributable to the random connectivity between nodes, which functions as implicit skip connections. These connections facilitate information flow and provide additional performance robustness compared to rigid chain-like structures.

\begin{table}[htbp]
  \centering
  \caption{Performance comparison between CogniSNN and the chain-like baseline on neuromorphic datasets and Tiny-ImageNet. Results are reported as mean $\pm$ standard deviation.}
  \label{tab:rga}
  \begin{tabular*}{\linewidth}{@{\extracolsep{\fill}}lccc}
    \toprule
    \textbf{Dataset} & \textbf{Baseline} & \textbf{\makecell[c]{CogniSNN\\(ER-RGA-7)}} & \textbf{\makecell[c]{CogniSNN\\(WS-RGA-7)}} \\
    \midrule
    DVS-Gesture   & 92.21$ \hspace{0.1em} \scriptstyle \pm \hspace{0.1em} 0.24$ & 94.78$ \hspace{0.1em} \scriptstyle \pm \hspace{0.1em} 0.12$ & \textbf{95.41}$ \hspace{0.1em} \scriptstyle \pm \hspace{0.1em} 0.21$ \\
    CIFAR10-DVS   & 77.12$ \hspace{0.1em} \scriptstyle \pm \hspace{0.1em} 0.15$ & \textbf{79.80}$ \hspace{0.1em} \scriptstyle \pm \hspace{0.1em} 0.10$ & 79.00$ \hspace{0.1em} \scriptstyle \pm \hspace{0.1em} 0.21$ \\
    N-Caltech101  & 76.57$ \hspace{0.1em} \scriptstyle \pm \hspace{0.1em} 0.20$ & \textbf{80.64}$ \hspace{0.1em} \scriptstyle \pm \hspace{0.1em} 0.15$ & 78.62$ \hspace{0.1em} \scriptstyle \pm \hspace{0.1em} 0.31$ \\
    Tiny-ImageNet & 52.50$ \hspace{0.1em} \scriptstyle \pm \hspace{0.1em} 0.20$ & \textbf{55.41}$ \hspace{0.1em} \scriptstyle \pm \hspace{0.1em} 0.17$ & 54.81$ \hspace{0.1em} \scriptstyle \pm \hspace{0.1em} 0.10$ \\
    \bottomrule
  \end{tabular*}
\end{table}

\begin{figure}[t]
    \centering
    \includegraphics[width=0.49\textwidth]{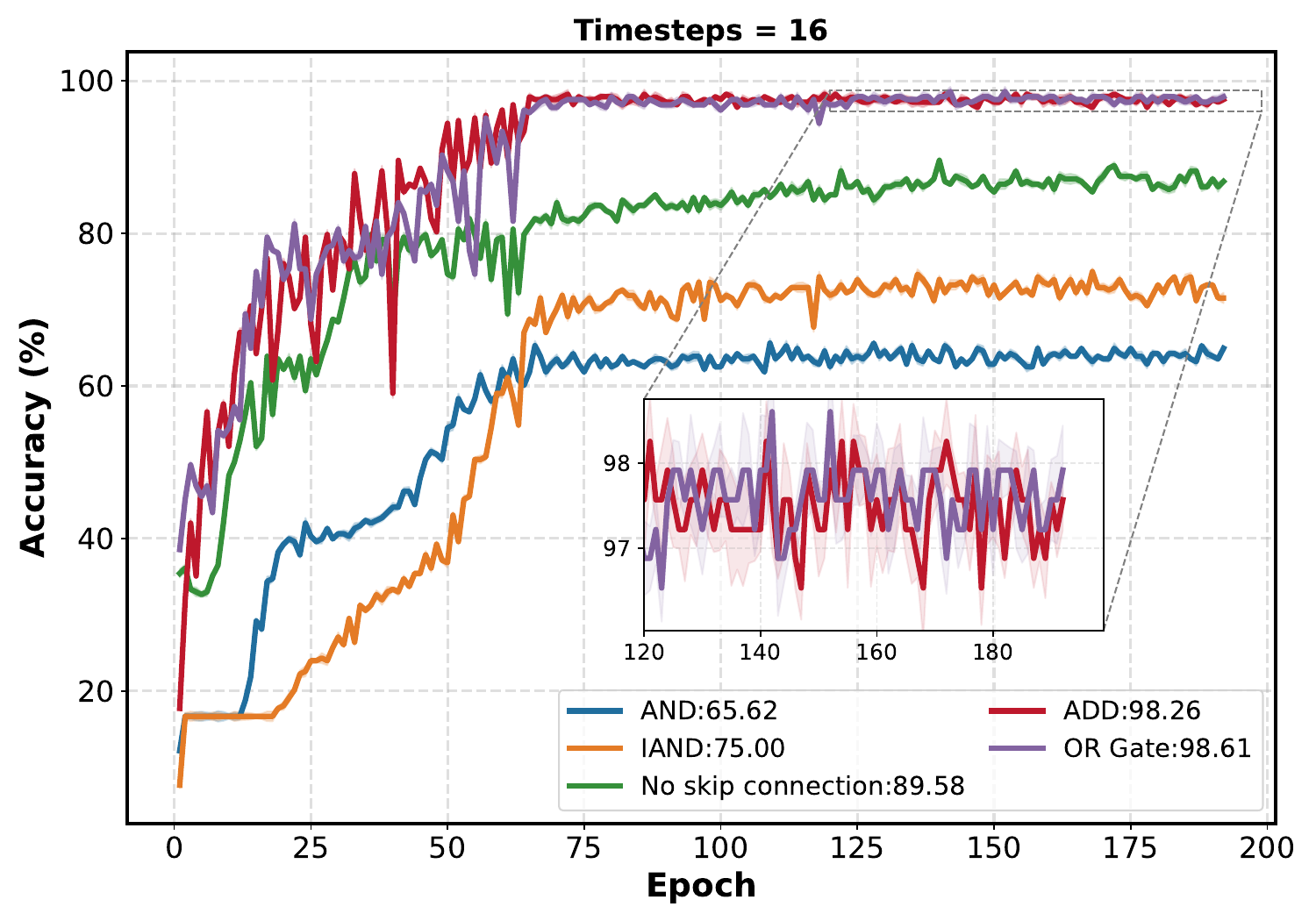}  
    \caption{Ablation study of residual mechanisms on the DVS-Gesture dataset with ER-RGA-7 based CogniSNN. ER-RGA-7 represents the ER-driven random graph architecture with 7 ResNodes.}
    \label{fig:ablation1-2}
\end{figure}
\textbf{2) Efficacy of the OR Gate:}
Subsequently, we evaluate the OR Gate against four alternative residual mechanisms using the ER-driven CogniSNN on the DVS-Gesture dataset at $T=16$. The comparison includes: (1) AND, (2) IAND, (3) No skip connection, (4) ADD, and (5) OR Gate (Ours). As shown in \textcolor{blue}{Fig.}~\ref{fig:ablation1-2}, the OR Gate yields a substantial accuracy improvement of $9.0\%$ over the ``No skip connection'' baseline, demonstrating its effectiveness in alleviating network degradation caused by deep neural pathways.

When compared with other logical operations, the AND mechanism is overly restrictive, emitting spikes only when both the residual and identity pathways fire simultaneously. Conversely, the IAND mechanism fails to emit spikes when both pathways fire, resulting in signal inhibition that contradicts biological intuition. Consequently, both mechanisms perform even worse than the baseline without skip connections.

Our OR Gate method also achieves a $0.4\%$ improvement over the ADD-based residual mechanism, yielding comparable performance while offering additional benefits. Unlike the ADD, which introduces floating-point computing and uncontrolled amplification of outputs, the OR Gate performs strictly binary operations. This characteristic ensures signal stability and highlights its potential as a robust and hardware-friendly spiking residual mechanism.

\subsection{Visualization Analysis}

\begin{figure}
	\centering
		\includegraphics[scale=0.3]{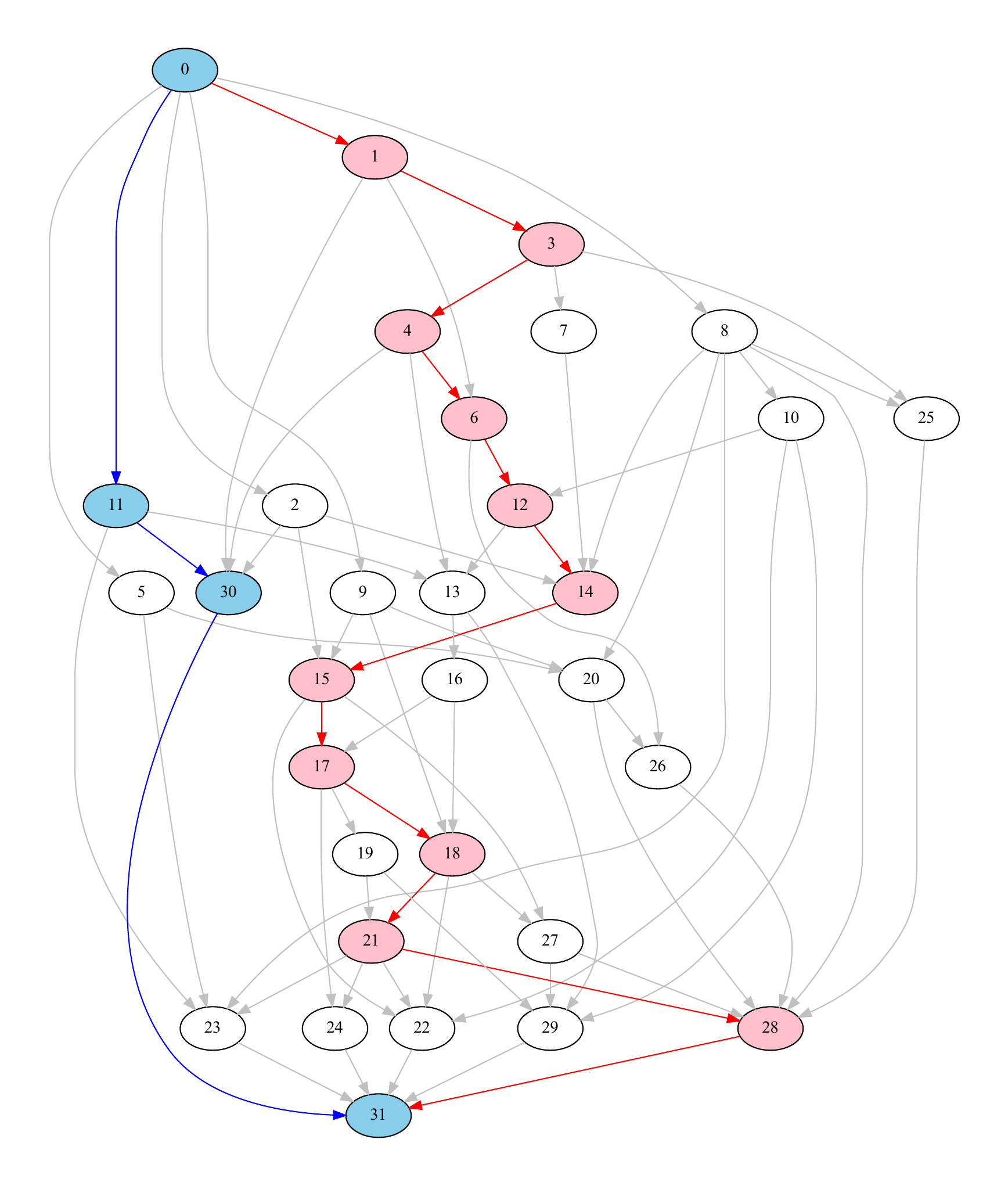}
	\caption{Visualization of the WS-driven Random Graph Architecture. The neural pathway with the highest BC is highlighted in pink, typically traversing the central region, while the pathway with the lowest BC is highlighted in blue, exhibiting a more peripheral structure.}
	\label{fig:WS32}
\end{figure}
\begin{figure*}[t]
    \centering
    \includegraphics[width=1\textwidth]{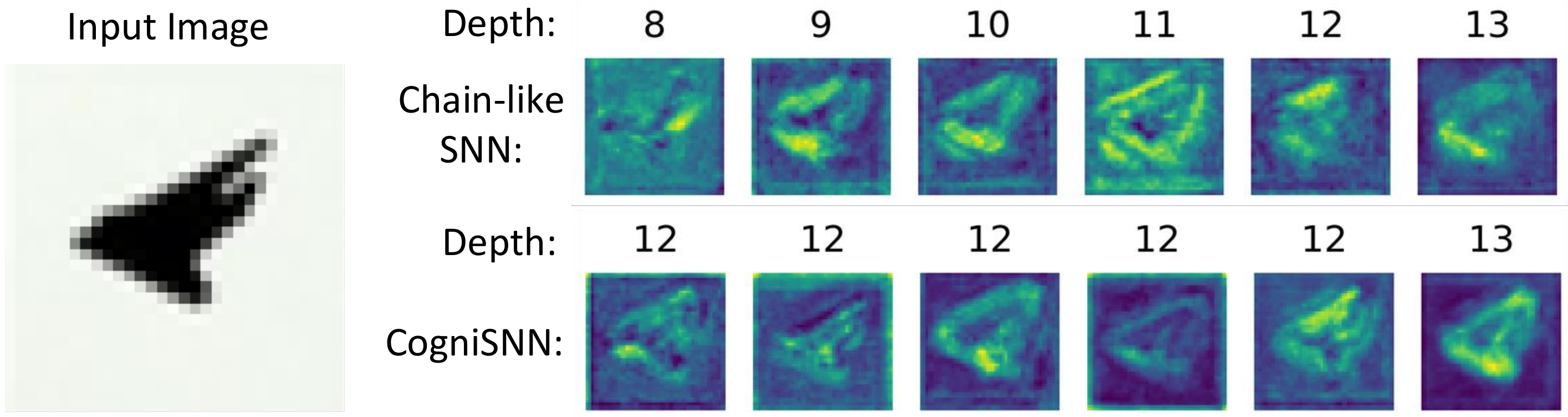}
    \caption{Visualization of feature maps comparing the Chain-like SNN and CogniSNN. Top: The chain-like structure shows feature maps from Depth 8 to 13. Bottom: CogniSNN displays outputs from five distinct penultimate nodes (all at Depth 12) and the final output node (Depth 13).}
    \label{fig: featureMap}
\end{figure*}

To further investigate the underlying reasons for the performance superiority of RGA-based SNNs over traditional chain-like architectures, we conduct a comparative analysis between a WS-driven CogniSNN (\textcolor{blue}{Fig.}~\ref{fig:WS32}) and a chain-like network comprising 13 nodes\footnote{The depth of 13 corresponds to the maximum depth of the WS-RGA-32 graph, ensuring a fair structural comparison based on network depth.}. Both networks are trained directly on the CIFAR10 dataset without spatial pooling. We subsequently visualize and compare the feature maps generated by the final layers of each architecture.

For CogniSNN, our visualization targets the output feature maps of the final node (Node 31) and its key predecessor nodes (specifically, Nodes 22, 23, 24, 28, and 29 in \textcolor{blue}{Fig.}~\ref{fig:WS32}). These nodes essentially function as the terminal points of distinct local pathways before final aggregation. Conversely, for the chain-like SNN, we visualize the outputs of the final six sequential nodes. The comparative results are presented in \textcolor{blue}{Fig.}~\ref{fig: featureMap}.

The visualizations demonstrate that CogniSNN captures significantly richer and more comprehensive feature representations at the final layer. In the chain-like SNN, feature quality stagnates or even deteriorates across the last six layers, indicative of representation degradation. In contrast, CogniSNN effectively leverages its graph topology: although individual penultimate nodes at depth 12 may not independently extract the strongest features, the final node at depth 13 integrates complementary signals from these diverse pathways to construct an optimal global representation. This multi-path integration mechanism is identified as the primary driver of its superior performance. Furthermore, the RGA topology introduces implicit skip connections, allowing valid low-level features to bypass intermediate processing and propagate directly to the output without degradation. This breaks the rigid paradigm of chain-like networks, where valuable low-level features are often diluted or compelled to transform into less relevant high-level abstractions.

Additionally, we visualize the neural pathways with the highest and lowest BC within the 32-node WS-driven RGA. As depicted in \textcolor{blue}{Fig.}~\ref{fig:WS32}, high-BC pathways are typically longer and densely distributed near the network's core, whereas low-BC pathways are shorter and located at the periphery. This structural observation reinforces our earlier conclusion: high-BC pathways are responsible for extracting robust, generalizable features suitable for transfer across similar scenarios, whereas low-BC pathways capture specialized or marginal features that are crucial for adapting to dissimilar tasks.

\subsection{Energy Analysis}

In this work, the integration of RGA into SNNs inevitably increases the computational load at the input stage of each node due to complex connectivity. To counterbalance this, we employ the pure spiking residual mechanism (OR Gate) to minimize floating-point operations. Additionally, the energy consumption of the DGL algorithm is unknown. Consequently, a detailed energy consumption analysis is essential.

\begin{table}[htbp]
  \centering
  \caption{The total energy consumption (mJ) for different tasks. The \textbf{bold} and \underline{underline} markers represent the best and second-best values.}
  \label{tab:energy_dynamic}
  
  \setlength{\tabcolsep}{2.5pt}

  \resizebox{\linewidth}{!}{
    \begin{tabular}{lccccc} 
      \toprule
      \multirow{2}{*}{\textbf{Dataset}} & \multirow{2}{*}{$T$} & 
      \textbf{\makecell{CogniSNN\\(OR)}} & 
      \textbf{\makecell{CogniSNN\\(ADD)}} & 
      \textbf{\makecell{Chain-like\\SNN (OR)}} & 
      \textbf{\makecell{CogniSNN\\(DGL)}} \\ 
      \midrule
      
      \multirow{3}{*}{DVS-Gesture} 
        & 5  & \textbf{7.46} & 8.89 & \underline{\textit{7.75}} & 8.19 \\ 
        & 8  & \textbf{7.86} & 9.02 & \underline{\textit{7.91}} & 8.77 \\
        & 16 & \underline{\textit{8.61}} & 9.72 & \textbf{8.43} & 9.31 \\
      \cmidrule{1-6} 
      
      \multirow{2}{*}{CIFAR10-DVS} 
        & 5  & \underline{\textit{67.21}} & 70.91 & \textbf{66.42} & 69.89 \\
        & 8  & \underline{\textit{58.29}} & 65.02 & \textbf{55.36} & 61.47 \\
      \cmidrule{1-6}
      
      \multirow{2}{*}{N-Caltech101} 
        & 5  & \underline{\textit{113.69}} & 115.94 & \textbf{110.88} & - \\
        & 8  & \underline{\textit{101.53}} & 114.15 & \textbf{98.56}  & - \\
      \bottomrule
    \end{tabular}%
  }
\end{table}
Specifically, we configure four distinct models for comparative analysis: (i) a 7-ResNode CogniSNN utilizing the OR mechanism; (ii) a 7-ResNode CogniSNN utilizing the ADD mechanism; (iii) a 7-ResNode Chain-like SNN utilizing the OR mechanism; and (iv) a CogniSNN with DGL, which incorporates the DGL algorithm into the first configuration.  In energy consumption simulation calculations, a single multiply and accumulation (MAC) operation consumes 4.6 pJ; a single accumulation (AC) operation consumes 0.9 pJ, as done in~\citet{2025_sun_ilif}.
Then, we evaluate their average energy consumption during inference across three neuromorphic datasets under different timestep settings.

As presented in \textcolor{blue}{Table}~\ref{tab:energy_dynamic}, under the identical RGA configuration, the OR mechanism consistently achieves higher energy efficiency than the ADD mechanism. This improvement stems from the binary nature of the OR operation, which substantially reduces the generation and accumulation of floating-point values. When comparing the RGA-based architecture with the chain-like structure, the chain-like SNN typically consumes less energy due to its sparser connectivity. Furthermore, regarding CogniSNN with DGL, we observe a marginal increase in energy consumption compared to the static CogniSNN. This is attributed to the fact that inference is performed on the full graph, and the DGL algorithm inherently encourages higher activation rates within specific subgraphs, leading to a slight uptick in total energy expenditure. Nevertheless, these marginal increases in energy cost are justifiable, as they are significantly outweighed by the substantial performance advantages of the CogniSNN framework, including its superior representational capacity, robustness, potential for continual learning, and deployment flexibility.

\section{Discussion}
\subsection{Difference from Network Architecture Search}

RGA-based NAS fundamentally operates as a search optimization process. It involves generating a multitude of random graphs by sampling a vast space of hyperparameters—such as generation protocols and edge connection probabilities—followed by extensive training to identify the single structure with optimal performance. In sharp contrast, our proposed RGA-based SNNs do not rely on such exhaustive searching. In our experiments, we utilize a fixed random seed to generate a single graph structure, and our results indicate that such generic random topologies consistently yield superior performance. To be precise, our reported accuracy is derived from multiple experiments on a fixed, randomly generated graph that was not selectively optimized. Unlike RGA-based NAS, which incurs significant computational costs to sample and train a vast number of candidates before identifying the optimal one, our approach demonstrates that the performance benefits stem from the intrinsic properties of the random topology itself, making the process significantly more time-efficient.

\subsection{Random Connections within Modules}

Stochastic connectivity serves as a critical factor in enhancing network robustness. Currently, RGA-based SNNs implement such random connections at the inter-node level (between basic ResNodes), allowing us to explore the macroscopic effects of node scale and neural pathway length across different orders of magnitude. Looking ahead, we aim to extend this paradigm to the intra-node level by introducing random connections directly between individual spiking neurons within layers. Furthermore, we plan to incorporate topological structures that dynamically evolve across the temporal domain, thereby simulating the brain's micro-level plasticity and structural fluidity.

\section{Conclusion}
In summary, this paper introduces a novel RGA paradigm for Spiking Neural Networks. Leveraging the rich neural pathways inherent in this topology, we explicitly model three brain-inspired properties: Neuron-Expandability, Pathway-Reusability, and Dynamic-Configurability. Collectively, these mechanisms endow the network with the capability to handle complex tasks, learn continually, and resist environmental interference. The experimental results are noteworthy. Neuron-Expandability, enabled by the pure spiking OR Gate and adaptive pooling, allows the network to match or even surpass current state-of-the-art performance. Pathway-Reusability, realized through the KP-LwF algorithm, underscores the importance of selectively recruiting specific neural pathways when addressing incremental learning across diverse scenarios. Dynamic-Configurability, supported by the Dynamic Growth Learning algorithm, not only significantly enhances robustness with reduced computational cost but also effectively mitigates the fixed-timestep constraints in real-world deployment. These advantages fully demonstrate the immense potential of SNNs with random graph architectures, paving the way for more robust and flexible brain-inspired computing systems.

\printcredits

\section*{Declaration of competing interest}
The authors declare that they have no known competing financial interests or personal relationships that could have appeared to influence the work reported in this paper.

\section*{AI declaration statement}
The authors utilized Gemini 3 Pro for language translation and polishing, reviewed the output, and take full responsibility for the content.

\section*{Acknowledgments}
This study is supported by Northeastern University, Shenyang, China (02110022124005), Government Special Support Funds for the Guangdong Institute of Intelligence Science and Technology, and National Natural Science Foundation of China (NSFC) under Grant No.62506084.

\section*{Data availability}
The code and datasets used in this study are available at \url{https://github.com/Yongsheng124/CogniSNN}. 

\bibliographystyle{cas-model2-names}

\bibliography{cas-refs}

@article{1977_freeman_set,
  title={A set of measures of centrality based on betweenness},
  author={Freeman, LC},
  journal={Sociometry},
  year={1977}
}

@article{2002_girvan_community,
  title={Community structure in social and biological networks},
  author={Girvan, Michelle and Newman, Mark EJ},
  journal={Proceedings of the national academy of sciences},
  volume={99},
  number={12},
  pages={7821--7826},
  year={2002},
  publisher={National Acad Sciences}
}

@inproceedings{2017_xie_aggregated,
  title={Aggregated residual transformations for deep neural networks},
  author={Xie, Saining and Girshick, Ross and Doll{\'a}r, Piotr and Tu, Zhuowen and He, Kaiming},
  booktitle={Proceedings of the IEEE conference on computer vision and pattern recognition},
  pages={1492--1500},
  year={2017}
}

@inproceedings{2019_xie_exploring,
  title={Exploring randomly wired neural networks for image recognition},
  author={Xie, Saining and Kirillov, Alexander and Girshick, Ross and He, Kaiming},
  booktitle={Proceedings of the IEEE/CVF International Conference on Computer Vision},
  pages={1284--1293},
  year={2019}
}

@article{2004_turing_intelligent,
  title={Intelligent machinery (1948)},
  author={Turing, Alan},
  journal={B. Jack Copeland},
  pages={395},
  year={1948}
}

@article{1958_rosenblatt_perceptron,
  title={The perceptron: a probabilistic model for information storage and organization in the brain.},
  author={Rosenblatt, Frank},
  journal={Psychological review},
  volume={65},
  number={6},
  pages={386},
  year={1958},
  publisher={American Psychological Association}
}

@article{1998_watts_collective,
  title={Collective dynamics of ‘small-world’networks},
  author={Watts, Duncan J and Strogatz, Steven H},
  journal={nature},
  volume={393},
  number={6684},
  pages={440--442},
  year={1998},
  publisher={Nature Publishing Group}
}

@article{2011_varshney_structural,
  title={Structural properties of the Caenorhabditis elegans neuronal network},
  author={Varshney, Lav R and Chen, Beth L and Paniagua, Eric and Hall, David H and Chklovskii, Dmitri B},
  journal={PLoS computational biology},
  volume={7},
  number={2},
  pages={e1001066},
  year={2011},
  publisher={Public Library of Science San Francisco, USA}
}

@article{2024_yan_sampling,
  title={Sampling complex topology structures for spiking neural networks},
  author={Yan, Shen and Meng, Qingyan and Xiao, Mingqing and Wang, Yisen and Lin, Zhouchen},
  journal={Neural Networks},
  volume={172},
  pages={106121},
  year={2024},
  publisher={Elsevier}
}

@inproceedings{2016_he_deep,
  title={Deep residual learning for image recognition},
  author={He, Kaiming and Zhang, Xiangyu and Ren, Shaoqing and Sun, Jian},
  booktitle={Proceedings of the IEEE conference on computer vision and pattern recognition},
  pages={770--778},
  year={2016}
}

@article{2021_hu_spiking,
  title={Spiking deep residual networks},
  author={Hu, Yangfan and Tang, Huajin and Pan, Gang},
  journal={IEEE Transactions on Neural Networks and Learning Systems},
  volume={34},
  number={8},
  pages={5200--5205},
  year={2021},
  publisher={IEEE}
}

@article{2021_fang_deep,
  title={Deep residual learning in spiking neural networks},
  author={Fang, Wei and Yu, Zhaofei and Chen, Yanqi and Huang, Tiejun and Masquelier, Timoth{\'e}e and Tian, Yonghong},
  journal={Advances in Neural Information Processing Systems},
  volume={34},
  pages={21056--21069},
  year={2021}
}

@article{2024_hu_advancing,
  title={Advancing spiking neural networks toward deep residual learning},
  author={Hu, Yifan and Deng, Lei and Wu, Yujie and Yao, Man and Li, Guoqi},
  journal={IEEE Transactions on Neural Networks and Learning Systems},
  year={2024},
  publisher={IEEE}
}

@article{2025_li_rethinking,
  title={Rethinking residual connection in training large-scale spiking neural networks},
  author={Li, Yudong and Lei, Yunlin and Yang, Xu},
  journal={Neurocomputing},
  volume={616},
  pages={128950},
  year={2025},
  publisher={Elsevier}
}

@article{2018_wu_spatio,
  title={Spatio-temporal backpropagation for training high-performance spiking neural networks},
  author={Wu, Yujie and Deng, Lei and Li, Guoqi and Zhu, Jun and Shi, Luping},
  journal={Frontiers in neuroscience},
  volume={12},
  pages={331},
  year={2018},
  publisher={Frontiers Media SA}
}

@article{2022_feng_multi,
  title={Multi-level firing with spiking ds-resnet: Enabling better and deeper directly-trained spiking neural networks},
  author={Feng, Lang and Liu, Qianhui and Tang, Huajin and Ma, De and Pan, Gang},
  journal={arXiv preprint arXiv:2210.06386},
  year={2022}
}

@inproceedings{2023_wang_spatial,
  title={Spatial-temporal self-attention for asynchronous spiking neural networks},
  author={Wang, Yuchen and Shi, Kexin and Lu, Chengzhuo and Liu, Yuguo and Zhang, Malu and Qu, Hong},
  booktitle={Proceedings of the Thirty-Second International Joint Conference on Artificial Intelligence},
  pages={3085--3093},
  year={2023}
}

@inproceedings{2024_lee_tt,
  title={TT-SNN: tensor train decomposition for efficient spiking neural network training},
  author={Lee, Donghyun and Yin, Ruokai and Kim, Youngeun and Moitra, Abhishek and Li, Yuhang and Panda, Priyadarshini},
  booktitle={2024 Design, Automation \& Test in Europe Conference \& Exhibition (DATE)},
  pages={1--6},
  year={2024},
  organization={IEEE}
}

@inproceedings{2022_li_neuromorphic,
  title={Neuromorphic data augmentation for training spiking neural networks},
  author={Li, Yuhang and Kim, Youngeun and Park, Hyoungseob and Geller, Tamar and Panda, Priyadarshini},
  booktitle={European Conference on Computer Vision},
  pages={631--649},
  year={2022},
  organization={Springer}
}

@article{2023_zhou_enhancing,
  title={Enhancing the performance of transformer-based spiking neural networks by SNN-optimized downsampling with precise gradient backpropagation},
  author={Zhou, Chenlin and Zhang, Han and Zhou, Zhaokun and Yu, Liutao and Ma, Zhengyu and Zhou, Huihui and Fan, Xiaopeng and Tian, Yonghong},
  journal={arXiv preprint arXiv:2305.05954},
  year={2023}
}

@inproceedings{2022_zhou_spikformer,
  title={Spikformer: When Spiking Neural Network Meets Transformer},
  author={Zhou, Zhaokun and Zhu, Yuesheng and He, Chao and Wang, Yaowei and Yan, Shuicheng and Tian, Yonghong and Yuan, Li},
  booktitle={The Eleventh International Conference on Learning Representations},
  year = {2023}
}

@inproceedings{2024_ding_shrinking,
  title={Shrinking Your TimeStep: Towards Low-Latency Neuromorphic Object Recognition with Spiking Neural Networks},
  author={Ding, Yongqi and Zuo, Lin and Jing, Mengmeng and He, Pei and Xiao, Yongjun},
  booktitle={Proceedings of the AAAI Conference on Artificial Intelligence},
  volume={38},
  pages={11811--11819},
  year={2024}
}

@inproceedings{2024_shen_tim,
  title={TIM: An efficient temporal interaction module for spiking transformer},
  author={Shen, Sicheng and Zhao, Dongcheng and Shen, Guobin and Zeng, Yi},
  booktitle={IJCAI},
  pages={3133--3141},
  year={2024}
}

@article{2023_guo_joint,
  title={Joint a-snn: Joint training of artificial and spiking neural networks via self-distillation and weight factorization},
  author={Guo, Yufei and Peng, Weihang and Chen, Yuanpei and Zhang, Liwen and Liu, Xiaode and Huang, Xuhui and Ma, Zhe},
  journal={Pattern Recognition},
  volume={142},
  pages={109639},
  year={2023},
  publisher={Elsevier}
}

@inproceedings{2022_kim_neural,
  title={Neural architecture search for spiking neural networks},
  author={Kim, Youngeun and Li, Yuhang and Park, Hyoungseob and Venkatesha, Yeshwanth and Panda, Priyadarshini},
  booktitle={European conference on computer vision},
  pages={36--56},
  year={2022},
  organization={Springer}
}

@article{2022_yang_training,
  title={Training spiking neural networks with local tandem learning},
  author={Yang, Qu and Wu, Jibin and Zhang, Malu and Chua, Yansong and Wang, Xinchao and Li, Haizhou},
  journal={Advances in Neural Information Processing Systems},
  volume={35},
  pages={12662--12676},
  year={2022}
}

@article{2020_deng_rethinking,
  title={Rethinking the performance comparison between SNNS and ANNS},
  author={Deng, Lei and Wu, Yujie and Hu, Xing and Liang, Ling and Ding, Yufei and Li, Guoqi and Zhao, Guangshe and Li, Peng and Xie, Yuan},
  journal={Neural networks},
  volume={121},
  pages={294--307},
  year={2020},
  publisher={Elsevier}
}

@article{2009_bullmore_complex,
  title={Complex brain networks: graph theoretical analysis of structural and functional systems},
  author={Bullmore, Ed and Sporns, Olaf},
  journal={Nature reviews neuroscience},
  volume={10},
  number={3},
  pages={186--198},
  year={2009},
  publisher={Nature Publishing Group UK London}
}

@article{2025_carboni_exploring,
  title={Exploring continual learning strategies in artificial neural networks through graph-based analysis of connectivity: Insights from a brain-inspired perspective},
  author={Carboni, Lucrezia and Nwaigwe, Dwight and Mainsant, Marion and Bayle, Rapha{\"e}l and Reyboz, Marina and Mermillod, Martial and Dojat, Michel and Achard, Sophie},
  journal={Neural Networks},
  volume={185},
  pages={107125},
  year={2025},
  publisher={Elsevier}
}

@article{2025_lu_estsformer,
  title={ESTSformer: Efficient spatio-temporal spiking transformer},
  author={Lu, Chengzhuo and Du, Huilin and Wei, Wenjie and Sun, Qian and Wang, Yuchen and Zeng, Dingyi and Chen, Wenyu and Zhang, Malu and Yang, Yang},
  journal={Neural Networks},
  pages={107786},
  year={2025},
  publisher={Elsevier}
}

@article{2025_zhang_combining,
  title={Combining aggregated attention and transformer architecture for accurate and efficient performance of Spiking Neural Networks},
  author={Zhang, Hangming and Sboev, Alexander and Rybka, Roman and Yu, Qiang},
  journal={Neural Networks},
  pages={107789},
  year={2025},
  publisher={Elsevier}
}

@article{2024_li_spikemba,
  title={Spikemba: Multi-modal spiking saliency mamba for temporal video grounding},
  author={Li, Wenrui and Hong, Xiaopeng and Xiong, Ruiqin and Fan, Xiaopeng},
  journal={arXiv preprint arXiv:2404.01174},
  year={2024}
}

@incollection{2020_tyagi_challenges,
  title={Challenges of applying deep learning in real-world applications},
  author={Tyagi, Amit Kumar and Rekha, Gillala},
  booktitle={Challenges and applications for implementing machine learning in computer vision},
  pages={92--118},
  year={2020},
  publisher={IGI Global Scientific Publishing}
}

@article{2012_dicarlo_does,
  title={How does the brain solve visual object recognition?},
  author={DiCarlo, James J and Zoccolan, Davide and Rust, Nicole C},
  journal={Neuron},
  volume={73},
  number={3},
  pages={415--434},
  year={2012},
  publisher={Elsevier}
}

@article{2023_han_adaptive,
  title={Adaptive reorganization of neural pathways for continual learning with spiking neural networks},
  author={Han, Bing and Zhao, Feifei and Pan, Wenxuan and Zhao, Zhaoya and Li, Xianqi and Kong, Qingqun and Zeng, Yi},
  journal={arXiv preprint arXiv:2309.09550},
  year={2023}
}

@article{2025_kudithipudi_neuromorphic,
  title={Neuromorphic computing at scale},
  author={Kudithipudi, Dhireesha and Schuman, Catherine and Vineyard, Craig M and Pandit, Tej and Merkel, Cory and Kubendran, Rajkumar and Aimone, James B and Orchard, Garrick and Mayr, Christian and Benosman, Ryad and others},
  journal={Nature},
  volume={637},
  number={8047},
  pages={801--812},
  year={2025},
  publisher={Nature Publishing Group UK London}
}

@inproceedings{2023_xu_unified,
  title={A unified structured framework for agi: Bridging cognition and neuromorphic computing},
  author={Xu, Mingkun and Zheng, Hao and Pei, Jing and Deng, Lei},
  booktitle={International Conference on Artificial General Intelligence},
  pages={345--356},
  year={2023},
  organization={Springer}
}

@inproceedings{2023_xu_exploiting,
  title={Exploiting homeostatic synaptic modulation in spiking neural networks for semi-supervised graph learning},
  author={Xu, Mingkun},
  booktitle={Proceedings of the 32nd ACM international conference on information and knowledge management},
  pages={5193--5195},
  year={2023}
}

@article{2021_xu_exploiting,
  title={Exploiting spiking dynamics with spatial-temporal feature normalization in graph learning},
  author={Xu, Mingkun and Wu, Yujie and Deng, Lei and Liu, Faqiang and Li, Guoqi and Pei, Jing},
  journal={arXiv preprint arXiv:2107.06865},
  year={2021}
}

@article{2024_xu_adaptive,
  title={Adaptive synaptic scaling in spiking networks for continual learning and enhanced robustness},
  author={Xu, Mingkun and Liu, Faqiang and Hu, Yifan and Li, Hongyi and Wei, Yuanyuan and Zhong, Shuai and Pei, Jing and Deng, Lei},
  journal={IEEE Transactions on Neural Networks and Learning Systems},
  volume={36},
  number={3},
  pages={5151--5165},
  year={2024},
  publisher={IEEE}
}

@article{2024_zhou_direct,
  title={Direct training high-performance deep spiking neural networks: a review of theories and methods},
  author={Zhou, Chenlin and Zhang, Han and Yu, Liutao and Ye, Yumin and Zhou, Zhaokun and Huang, Liwei and Ma, Zhengyu and Fan, Xiaopeng and Zhou, Huihui and Tian, Yonghong},
  journal={Frontiers in Neuroscience},
  volume={18},
  pages={1383844},
  year={2024},
  publisher={Frontiers Media SA}
}

@article{2025_shen_context,
  title={Context gating in spiking neural networks: Achieving lifelong learning through integration of local and global plasticity},
  author={Shen, Jiangrong and Ni, Wenyao and Xu, Qi and Pan, Gang and Tang, Huajin},
  journal={Knowledge-Based Systems},
  volume={311},
  pages={112999},
  year={2025},
  publisher={Elsevier}
}

@article{2023_han_enhancing,
  title={Enhancing efficient continual learning with dynamic structure development of spiking neural networks},
  author={Han, Bing and Zhao, Feifei and Zeng, Yi and Pan, Wenxuan and Shen, Guobin},
  journal={arXiv preprint arXiv:2308.04749},
  year={2023}
}

@article{2017_li_learning,
  title={Learning without forgetting},
  author={Li, Zhizhong and Hoiem, Derek},
  journal={IEEE transactions on pattern analysis and machine intelligence},
  volume={40},
  number={12},
  pages={2935--2947},
  year={2017},
  publisher={IEEE}
}

@article{2025_ma_efficient,
  title={Efficient Training of Spiking Neural Networks by Spike-aware Data Pruning},
  author={Ma, Chenxiang and Chen, Xinyi and Wu, Yujie and Tan, Kay Chen and Wu, Jibin},
  journal={arXiv preprint arXiv:2510.04098},
  year={2025}
}

@article{2025_huang_cognisnn,
  title={CogniSNN: A First Exploration to Random Graph Architecture based Spiking Neural Networks with Enhanced Expandability and Neuroplasticity},
  author={Huang, Yongsheng and Duan, Peibo and Liu, Zhipeng and Sun, Kai and Zhang, Changsheng and Zhang, Bin and Xu, Mingkun},
  journal={arXiv preprint arXiv:2505.05992},
  year={2025}
}

@article{2002_werbos_backpropagation,
  title={Backpropagation through time: what it does and how to do it},
  author={Werbos, Paul J},
  journal={Proceedings of the IEEE},
  volume={78},
  number={10},
  pages={1550--1560},
  year={2002},
  publisher={IEEE}
}

@article{1994_white_betweenness,
  title={Betweenness centrality measures for directed graphs},
  author={White, Douglas R and Borgatti, Stephen P},
  journal={Social networks},
  volume={16},
  number={4},
  pages={335--346},
  year={1994},
  publisher={Elsevier}
}

@article{2025_yu_efficient,
  title={Efficient Logit-based Knowledge Distillation of Deep Spiking Neural Networks for Full-Range Timestep Deployment},
  author={Yu, Chengting and Zhao, Xiaochen and Liu, Lei and Yang, Shu and Wang, Gaoang and Li, Erping and Wang, Aili},
  journal={arXiv preprint arXiv:2501.15925},
  year={2025}
}

@article{2025_sun_ilif,
  title={ILIF: Temporal Inhibitory Leaky Integrate-and-Fire Neuron for Overactivation in Spiking Neural Networks},
  author={Sun, Kai and Duan, Peibo and Kuhlmann, Levin and Wang, Beilun and Zhang, Bin},
  journal={arXiv preprint arXiv:2505.10371},
  year={2025}
}

@inproceedings{2017_amir_low,
  title={A low power, fully event-based gesture recognition system},
  author={Amir, Arnon and Taba, Brian and Berg, David and Melano, Timothy and McKinstry, Jeffrey and Di Nolfo, Carmelo and Nayak, Tapan and Andreopoulos, Alexander and Garreau, Guillaume and Mendoza, Marcela and others},
  booktitle={Proceedings of the IEEE conference on computer vision and pattern recognition},
  pages={7243--7252},
  year={2017}
}

@article{2017_li_cifar10,
  title={Cifar10-dvs: an event-stream dataset for object classification},
  author={Li, Hongmin and Liu, Hanchao and Ji, Xiangyang and Li, Guoqi and Shi, Luping},
  journal={Frontiers in neuroscience},
  year={2017},
  publisher={Frontiers Media SA}
}

@article{2015_orchard_converting,
  title={Converting static image datasets to spiking neuromorphic datasets using saccades},
  author={Orchard, Garrick and Jayawant, Ajinkya and Cohen, Gregory K and Thakor, Nitish},
  journal={Frontiers in neuroscience},
  volume={9},
  pages={437},
  year={2015},
  publisher={Frontiers Media SA}
}

@article{2015_le_tiny,
  title={Tiny imagenet visual recognition challenge},
  author={Le, Yann and Yang, Xuan},
  journal={CS 231N},
  volume={7},
  number={7},
  pages={3},
  year={2015}
}

@article{2017_heusel_gans,
  title={Gans trained by a two time-scale update rule converge to a local nash equilibrium},
  author={Heusel, Martin and Ramsauer, Hubert and Unterthiner, Thomas and Nessler, Bernhard and Hochreiter, Sepp},
  journal={Advances in neural information processing systems},
  volume={30},
  year={2017}
}

@inproceedings{2019_wu_direct,
  title={Direct training for spiking neural networks: Faster, larger, better},
  author={Wu, Yujie and Deng, Lei and Li, Guoqi and Zhu, Jun and Xie, Yuan and Shi, Luping},
  booktitle={Proceedings of the AAAI conference on artificial intelligence},
  volume={33},
  number={01},
  pages={1311--1318},
  year={2019}
}

@article{2025_shi_hybrid,
  title={Hybrid neural networks for continual learning inspired by corticohippocampal circuits},
  author={Shi, Qianqian and Liu, Faqiang and Li, Hongyi and Li, Guangyu and Shi, Luping and Zhao, Rong},
  journal={Nature Communications},
  volume={16},
  number={1},
  pages={1272},
  year={2025},
  publisher={Nature Publishing Group UK London}
}

@inproceedings{2021_zheng_going,
  title={Going deeper with directly-trained larger spiking neural networks},
  author={Zheng, Hanle and Wu, Yujie and Deng, Lei and Hu, Yifan and Li, Guoqi},
  booktitle={Proceedings of the AAAI conference on artificial intelligence},
  volume={35},
  number={12},
  pages={11062--11070},
  year={2021}
}

@article{2020_he_comparing,
  title={Comparing SNNs and RNNs on neuromorphic vision datasets: Similarities and differences},
  author={He, Weihua and Wu, YuJie and Deng, Lei and Li, Guoqi and Wang, Haoyu and Tian, Yang and Ding, Wei and Wang, Wenhui and Xie, Yuan},
  journal={Neural Networks},
  volume={132},
  pages={108--120},
  year={2020},
  publisher={Elsevier}
}

@article{2020_deng_tianjic,
  title={Tianjic: A unified and scalable chip bridging spike-based and continuous neural computation},
  author={Deng, Lei and Wang, Guanrui and Li, Guoqi and Li, Shuangchen and Liang, Ling and Zhu, Maohua and Wu, Yujie and Yang, Zheyu and Zou, Zhe and Pei, Jing and others},
  journal={IEEE Journal of Solid-State Circuits},
  volume={55},
  number={8},
  pages={2228--2246},
  year={2020},
  publisher={IEEE}
}

@article{2022_wu_brain,
  title={Brain-inspired global-local learning incorporated with neuromorphic computing},
  author={Wu, Yujie and Zhao, Rong and Zhu, Jun and Chen, Feng and Xu, Mingkun and Li, Guoqi and Song, Sen and Deng, Lei and Wang, Guanrui and Zheng, Hao and others},
  journal={Nature Communications},
  volume={13},
  number={1},
  pages={65},
  year={2022},
  publisher={Nature Publishing Group UK London}
}

\end{document}